\documentclass[journal]{IEEEtran}

\usepackage[numbers,sort&compress]{natbib}
\usepackage{amsmath,epsfig,amssymb,booktabs,multirow,makecell,subfigure,graphicx,diagbox}
\usepackage{enumitem}
\usepackage{soul} 
\usepackage{color, xcolor} 

\usepackage[ruled, lined, linesnumbered, commentsnumbered, longend]{algorithm2e}
\usepackage{algorithmic}

\usepackage{threeparttable}
\usepackage{setspace}
\usepackage{soul}
\usepackage{bbm}
\usepackage{gensymb}
\usepackage{pifont}

\usepackage{xcolor}
\usepackage{xpatch}

\makeatletter
\renewcommand{\maketag@@@}[1]{\hbox{\m@th\normalsize\normalfont#1}}%
\makeatother

\begin{document}
\title{
ECP-Mamba: An Efficient Multi-scale Self-supervised Contrastive Learning Method 
with State Space Model
for PolSAR Image Classification}
\author{        
        Zuzheng Kuang, 
        Haixia Bi, 
        Chen Xu, 
        Jian Sun                 
        
        \thanks{This paper was supported by the National Key R\&D Program of China under Grant 2022YFA1003800, NSFC under Grant 42201394 and 12326615, and Qinchuangyuan Talent Program under grant QCYRCXM-2022-30. 
        
Haixia Bi, Zuzheng Kuang and Fan Li are with School of Information and Communications Engineering, Xi'an Jiaotong University, Xi’an 710049, China;
(mail:{haixia.bi@xjtu.edu.cn, kzz794466014@stu.xjtu.edu.cn})

Chen Xu is with Department of Mathematics and Fundamental Research,  
Peng Cheng Laboratory, Shenzhen 518055, China
and School of Mathematics and Statistics, Xi’an Jiaotong University,
Xi’an 710049, China; (mail: xuch01@pcl.ac.cn)

Jian Sun is with School of Mathematics and Statistics, Xi’an Jiaotong University, Xi’an 710049, China.
(mail:{jiansun@xjtu.edu.cn.})}}

\markboth{IEEE TRANSACTIONS on GEOSCIENCE AND REMOTE SENSING, June~2025}%
{Shell \MakeLowercase{\textit{et al.}}: }

\maketitle

\begin{abstract}
Recently, 
polarimetric synthetic aperture radar (PolSAR) image classification
has been greatly promoted by 
deep neural networks. 
However,
current deep learning-based PolSAR classification methods 
encounter difficulties 
due to its dependence on extensive labeled data and
the computational inefficiency of architectures like Transformers. 
This paper presents ECP-Mamba, 
an efficient framework integrating multi-scale self-supervised contrastive learning with a state space model (SSM) backbone. 
Specifically, 
ECP-Mamba addresses annotation scarcity
through a multi-scale predictive pretext task 
based on local-to-global feature correspondences,
which uses a simplified self-distillation paradigm without negative sample pairs.
To enhance computational efficiency,
the Mamba architecture (a selective SSM) is first tailored for pixel-wise PolSAR classification task 
by designing a spiral scan strategy. 
This strategy prioritizes causally relevant features near the central pixel, 
leveraging the localized nature of pixel-wise classification tasks.
Additionally, 
the lightweight Cross Mamba module is proposed to facilitates complementary multi-scale feature interaction 
with minimal overhead. 
Extensive experiments across four benchmark datasets
demonstrate ECP-Mamba's effectiveness in
balancing high accuracy with resource efficiency. 
On the Flevoland 1989 dataset, 
ECP-Mamba achieves state-of-the-art performance with 
an overall accuracy of 99.70\%, average accuracy of 99.64\% and Kappa coefficient of 99.62e-2.
Our code will be available at
https://github.com/HaixiaBi1982/ECP\_Mamba.

\end{abstract}

\IEEEpeerreviewmaketitle

\section{Introduction}

\IEEEPARstart{T}{he} advent of Polarimetric Synthetic Aperture Radar (PolSAR) technology has revolutionized the field of remote sensing by significantly enhancing our ability to observe and interpret the Earth surface~\cite{parikh2020classification}.
Unlike SAR systems that operate with a single polarization, PolSAR systems employ all four possible polarizations (HH, HV, VH and VV), offering a wealth of information about imaged targets’ scattering characteristics~\cite{han2023polarimetric}.
Due to its comprehensive polarization imaging and high penetrative abilities, 
PolSAR enables more reliable estimation, classification, and verification of the observed objects 
under any time and any weather conditions~\cite{bi2017unsupervised}. 
This progressive technology provides persistent monitoring capabilities, 
rendering it a promising tool for remote sensing applications.
%
As an advanced and widely interested task of PolSAR realm, 
PolSAR image classification, 
which entails the allocation of landscape categories to each pixel,
has broadened the applicability of PolSAR across a spectrum of domains from
agricultural yield prediction~\cite{kuang2024polarimetry}
to disaster recovery~\cite{yang2024meta}.

The PolSAR image classification task has been extensively explored,
producing numerous methods such as
target decomposition-based,
statistical distribution-based, 
and machine learning-based approaches~\cite{bi2020polarimetric}.
Target decomposition-based methods 
decompose scattering matrices into components with specific physical interpretations, 
such as 
Pauli decomposition~\cite{cloude1996review},
Freeman decomposition~\cite{freeman1998three} and 
Yamaguchi decomposition~\cite{yamaguchi2005four} etc.
Statistical distribution-based methods
are predominantly based on statistical models,
such as
${K}$-distribution~\cite{lee1994k}, 
Wishart distribution~\cite{lee1999unsupervised} and 
$\mathcal{U}$-distribution~\cite{doulgeris2008classification} etc.
During the past decade, 
machine learning-based methods have become prominent
with the implementation of
Support Vector Machine~\cite{SVM_in_PolSAR2009},
K-Nearest Neighbors~\cite{k_nearest_neighbor2015}
and Bayesian learning~\cite{9048587} etc.
Different from the aforementioned methods relying on 
hand-crafted features,
deep learning (DL)-based approaches offer an end-to-end learning framework,
which is characterized by their automatic feature extraction from PolSAR data 
and powerful capacity for encoding information~\cite{kuang2024polarimetry}.
The development of DL models has considerably advanced the PolSAR image classification performance, 
capturing the complexity and variability of real-world scenarios with improved results
\cite{zhouyuCNN, 
baseline_of_fudan_CV_CNN_2017,
wang2018exploring, 
PCN,
dong2021exploring, 
jamali2023local, 
geng2024polarimetric}.
Nonetheless, 
these supervised approaches have 
a considerable reliance on datasets with extensive volumes of annotations
and disregard for the high computational demands, 
which pose two primary issues. 
Firstly, 
obtaining ample annotated datasets for PolSAR data is laborious 
due to the specialized expertise required for accurate annotation.
%
Secondly, training on vast datasets often demands substantial computational resources and time, 
which can be prohibitive for many applications.
%
Therefore, 
to achieve a commendable balance of performance and efficiency,
it is imperative to develop 
DL-based PolSAR image classification methods 
with \textbf{scare annotation adaptability} and \textbf{computational effectiveness}.

Self-supervised learning (SSL) develops features independently of manual labels, 
which is a powerful alternative to limited label scenarios.
Based on pretext task designs, 
SSL can be primarily categorized into
generative learning (GL) and 
contrastive learning (CL)
\cite{kuang2024polarimetry}.
GL-based methods
\cite{SS_CV-GAN_4_PolSAR, wang2024dual, Kuang2025CVPolDiff}
learn features through a process that 
involves recovering data from its corrupted forms and 
addressing challenges encountered during the generation process,
which have been widely studied nowadays.
CL-based methods
\cite{first_explore_in_self-supervised_2020,
Mutual_Information_Based_Self-Supervised_2021,
BYOL_in_PolSAR_Images_2022,
kuang2024polarimetry}
learn features by contrasting augmented data views,
focusing on the similarity of same-sample pairs (positive pairs)
and the differentiation of dissimilar pairs (negative pairs).
However, 
existing CL-based PolSAR image classification researches 
mainly concentrated on
instance-wise distinctions, 
overlooking the essential 
scale-wise consistency and complementary 
crucial for capturing data complexity 
through multi-scale feature learning.
%
Taking \emph{water} class in the AIRSAR Flevoland dataset~\cite{graph-Bi2018} for example.
Both its local and global features share similarities,
such as the basic annotation properties,
while the former focuses on fine-grained information like scattering intensity; 
the latter is conducive to understanding the contextual structure within the image, 
such as the spatial relationship between water and adjacent land. 
Therefore,
contrastive learning local and global views of one target
not only ensures the consistency of multi-scale features, 
but also enables complementary PolSAR data modeling
from various perspectives.
%
Powered by this,
we introduce a scale-invariant CL framework 
incorporating knowledge distillation 
to learn multi-scale representations 
from local-to-global correspondences.

How to find an excellent network architecture has been a focal point of DL research~\cite{malhotra2023recent}.
Different network structures essentially impact the model's learning capability, computational intensity, and 
potential scalability.
%
Convolutional Neural Networks (CNNs) have been well-developed
and successfully applied in PolSAR Image Classification
\cite{zhouyuCNN, 
baseline_of_fudan_CV_CNN_2017, 
parikh2023modeling, 
kuang2024polarimetry, 
Kuang2025CVPolDiff}.
However, their inherited shortfall lies in 
the incapacity to comprehensively obtain responses from informative regions throughout the entire input, 
offering suboptimal results to global feature analysis
\cite{zhu2024vision}.
%
Transformers overcome this by employing attention mechanism
\cite{vaswani2017attention}.
This architecture excels at capturing long-range dependencies and
extracting comprehensive contextual cues, 
thereby delivering outstanding classification results
\cite{dong2021exploring, jamali2023local}.
Despite the robust representational capability, 
Transformer-based models still confront expensive computational challenges in real-world applications, 
whose self-attention mechanism incurs quadratic computational complexity $\mathcal{O}(N^{2})$ with input size $N$,
undermining models' efficiency
in terms of 
floating point operations (FLOPs)
\cite{gu2023mamba, zhu2024vision}.
Fortunately, State Space Models (SSMs), 
particularly the structured SSM (S4) \cite{gu2021efficiently}, 
have emerged as a promising alternative to Transformer architecture. 
Through integrating S4 with deep learning theory, 
the selective structured state space model (Mamba) \cite{gu2023mamba}
has been introduced as the advanced architecture
which not only achieves near-linear complexity and 
long-range dependencies,
but also demonstrates superior performance across multiple tasks \cite{xu2024visualmambasurveynew}.
Although SSMs have been thoroughly applied to 
computer vision
\cite{zhu2024vision}, 
their implementations in PolSAR remain largely unexplored.
Driven by this,
we are dedicated to making the first attempt to explore the potential of SSMs 
for PolSAR image classification.
%

Overall, we introduce ECP-Mamba, 
an efficient multi-scale self-supervised contrastive
learning method with state space model 
for PolSAR image classification.
The key contributions of our study are as follows:

\begin{itemize}
\item [1)] 

To address the scarce annotation problem,
we present a self-distillation pattern 
designed to exploit the consistent and complementary 
multi-scale features 
across local-to-global views.
%

\begin{itemize}
\item[$\bullet$] 
By leveraging CL on features of the same scattering object across different scales, 
the model's robustness to varying scales and classes is simultaneously enhanced.

\item[$\bullet$] 
Independent of 
complex momentum encoders,
negative sample pairs,
and large backbone networks,
the self-distillation pattern simplifies implementation and enhances scalability.

\end{itemize}

\item [2)]
Addressing computational inefficiencies without compromising accuracy,
we take the first step to investigate the adaptation
of SSMs in PolSAR image classification
and propose Multi-scale Efficient Mamba.

\begin{itemize}
\item[$\bullet$] 
We customize a Mamba-based block 
for pixel-wise classification task.
We demonstrate that pixels closer to the central pixel possess stronger causal relationships with annotation,
and design a spiral scanning technique within the Mamba architecture.

\item[$\bullet$] 
To encourage the complementary between 
multi-scale features, 
we newly devise the Cross Mamba module 
to further augment the interaction of local and global features
without imposing extra computational demands.
\end{itemize}

\end{itemize}

Our research involves 4 benchmark datasets. 
Through comparative experiments
against state-of-the-art methods,
results and analyses clearly establish the superiority of 
our model's capability and efficiency.

The paper structure is outlined below: 
Section \ref{Related_Works} offers a concise overview of the related works.
Section \ref{Overview} to Section \ref{Summary} provide detailed descriptions of the proposed methodology.
Section \ref{Experiments} details the experiments and analysis. 
Section \ref{Conclusion} summarizes the conclusion and suggests future directions.

\section{Related Works}\label{Related_Works}
\subsection{Imaging of PolSAR data}

Using the vector characteristics of electromagnetic waves, 
PolSAR systems obtain backscattered waves by alternately transmitting and simultaneously receiving electromagnetic waves in different polarization modes. 
This unique imaging mechanism leads to the complex-valued format of PolSAR data, 
which contains both amplitude and phase information.
%
%
The scattering characteristics of targets can be represented by a $2 \times 2$ complex polarimetric scattering matrix $\mathbf{S}$~\cite{polarimetric_scattering_matrix1994}:
\begin{equation}
\mathbf{S}=\left[\begin{array}{ll}
S_{\mathbf{H H}} & S_{\mathbf{H V}} \\
S_{\mathbf{V H}} & S_{\mathbf{V V}}
\end{array}\right],
\end{equation}
where $S_{pq}$ indicates the scattering element with incident polarization $p$ 
and receiving polarization $q$,
while $\mathbf{H}$ and $\mathbf{V}$ signify the horizontal and vertical polarization modes respectively.
%
%
Under the reciprocity theorem,  
i.e., $S_{\mathbf{H V}} = S_{\mathbf{V H}}$,
scattering matrix $\mathbf{S}$ can be simplified into a 3 dimensional (3D) scattering vector $\mathbf{p}$  
by Pauli decomposition~\cite{Pauli_decomposition_of_scattering_matrix1996}: 
\begin{equation}
\mathbf{p}=\frac{1}{\sqrt{2}}\left[S_{\mathbf{H H}}+S_{\mathbf{V V}} \quad S_{\mathbf{H H}}-S_{\mathbf{V V}} \quad 2S_{\mathbf{H V}}\right]^{T},
\end{equation}
where $^{T}$ represents transposition operation. 
For the multi-look case, 
the second-order spatial averaging coherence matrix $\mathbf{T}$ 
can be represented as~\cite{baseline_of_fudan_CV_CNN_2017}: 
%
%
\begin{equation}\label{T}
\mathbf{T}=\frac{1}{L_s} 
\sum_{i=1}^{L_s} \mathbf{p}_{i} \mathbf{p}_{i}^{H}=\left[\begin{array}{ccc}
T_{11} & T_{12} & T_{13} \\
T_{21} & T_{22} & T_{23} \\
T_{31} & T_{32} & T_{33}
\end{array}\right].
\end{equation}
Here, $L_s$ is the number of looks and $^{H}$ stands for conjugate transposition.
It should be noted that $\mathbf{T}$ is a $3 \times 3$ complex conjugate symmetric matrix
with real-valued diagonal elements and complex-valued off-diagonal elements.

\subsection{Deep learning to Contrastive learning in PolSAR}

DL architectures including
CNNs~\cite{zhouyuCNN, baseline_of_fudan_CV_CNN_2017},
Recurrent Neural Networks (RNNs)~\cite{wang2018exploring, ni2020random}
and Vision Transformers (ViTs)~\cite{dong2021exploring, jamali2023local, geng2024polarimetric}
have demonstrated superior results in PolSAR image classification. 
However, 
these supervised studies typically require extensive labeled datasets for training.
which may suffer from performance collapse and overfitting
in scare annotation environments
\cite{graph-Bi2018,first_Active_learning_2019}.

Contrastive learning, 
an effective SSL path to solve this challenge,
has delivered remarkable results in the computer vision field
\cite{InstDisc, Moco, InfoMin, BYOL, DINO}.
CL extracts discriminative features through 
harnessing the intrinsic differentiation within unlabeled data.
Wu et al.~\cite{InstDisc} pioneered this concept
by presenting a pretext task of instance discrimination.
%
Inspired by this,
subsequent studies~\cite{Moco, InfoMin} focused on refining the model's ability to discern between positive and negative sample pairs through pretext tasks.
In contrast, literature~\cite{BYOL, DINO} argued that predicting one sample from itself without negative pairs can simplify the contrastive process with comparable performance.
%
In the context of PolSAR image classification, 
CL-based methods
\cite{first_explore_in_self-supervised_2020, Mutual_Information_Based_Self-Supervised_2021, BYOL_in_PolSAR_Images_2022, hua2024multi, kuang2024polarimetry}
mainly emerged in recent years.
Zhang et al.~\cite{first_explore_in_self-supervised_2020} first used CL in PolSAR 
with an instance-level discrimination task and a memory bank for feature recording.
Ren et al.~\cite{Mutual_Information_Based_Self-Supervised_2021} introduced a CNN-based method 
to extract mutual information by discerning different modalities of PolSAR instances.
%
Zhang et al.~\cite{BYOL_in_PolSAR_Images_2022} 
designed a three-stream CNN framework for negative sample-free CL by generating positive samples.
%
%
%
Hua et al.~\cite{hua2024multi} proposed a multi-scale CL model,
whose augmentation module involved random crop
and networks integrated multiple sizes of 
convolutional heads for multi-scale feature extraction.
Inspired by the prior knowledge of PolSAR,
Kuang et al.~\cite{kuang2024polarimetry}
developed a complex-valued CL method for 
class-imbalanced classification.

\subsection{State Space Model and Its Applications in Remote Sensing}
As a versatile concept across disciplines,
SSM is crucial for connecting input and output sequences through latent states,
aiding sequence transformation 
within deep neural networks~\cite{xu2024visualmambasurveynew}.
Early SSM-based deep learning faced computational challenges, which were addressed by S4 \cite{gu2021efficiently} 
through reparameterizing state matrices
and enhancing memory efficiency.
Mamba \cite{gu2023mamba} is one of the landmark works in S4. 
It introduces the selection mechanism to 
address the limitation of constant sequence transitions, 
enabling selective information propagation or forgetting based on the current context.
This innovation allows Mamba to compete with Transformers in modeling capabilities while ensuring linear scalability, 
which represents a significant advancement in long-sequences processing~\cite{xu2024visualmambasurveynew}.
The evolution in SSMs has accelerated the development of visual Mamba models. 
Forerunners like Vim~\cite{zhu2024vision} and VMamba~\cite{liu2024vmambavisualstatespace}
have proved the feasibility of 
managing image sequences via 
integrating spatial information and bi-directional SSMs,
showcasing Mamba's adaptability and effectiveness 
in computer vision.
To adapt Mamba for different tasks, 
diverse scanning strategies have been crafted to
tackle the challenges associated 
with non-causal image data~\cite{liu2024vision}.
Techniques like Zigzag Scan~\cite{yang2024plainmamba}, 
Local Scan~\cite{huang2024localmamba}, 
Hilbert Scan~\cite{he2024mambaad} 
and Atrous Scan~\cite{pei2024efficientvmamba} 
have proven to be effective.
%
%

In the last year, 
there have been several works in integrating SSMs into remote sensing tasks, such as 
image classification~\cite{chen2024rsmamba, yao2024spectralmamba}, 
image denoising~\cite{fu2024ssumamba}
and change detection~\cite{chen2024changemamba}.
%
Chen et al.~\cite{chen2024rsmamba} improved classification results
by using a dynamic multi-scanning mechanism with shuffle scan for SSMs.
%
In spectral image classification, 
Yao et al.~\cite{yao2024spectralmamba} introduced 
an efficient framework
using a piece-wise scanning mechanism
and a spatial-spectral merging block
within downsizing SSMs.
%
In hyperspectral image denoising,
Fu et al.~\cite{fu2024ssumamba} proposed a spatial-spectral SSM 
incorporating 3D CNN
that alternates scanning through rows, columns and bands in 6 directions. 
%
Chen et al.~\cite{chen2024changemamba} explored
a SSM-based model for remote sensing change detection,
focusing on spatio-temporal data modeling and
multi-level features fusion.

Based on the above-mentioned studies, 
the following conclusions can be reached:

\begin{itemize}
\item[1)]
Existing CL-based PolSAR image classification works focus on
capturing differences among instances across augmented views.
However, the potential of multi-scale features in CL remains under-explored.
Although \cite{hua2024multi} mentioned multi-scale CL,
its random cropping leads to 
an inconsistent extraction of multi-scale information.
Furthermore, employing branches of convolutions 
could introduce inductive bias and information redundancy.

\item[2)]
Recent PolSAR image classification methods 
tend to rely on Transformer,
which overlook the resource-intensity when 
network deepens or practical training.

\item[3)]
The prominence of SMMs
underline the importance of 
tailoring scanning strategy when 
adapting Mamba to different downstream tasks, 
which still remains an uncharted territory
for pixel-level PolSAR image classification.

\item[4)]
Most CL works 
require one or more of the following conditions:
complex momentum encoders,
negative sample pairs
or high computational complexity.
Contrastively, 
designing an efficient CL framework 
presents a scientifically meritorious yet unrealized topic.

\end{itemize}

The unique attributes of our proposed approach compared to existing research lie in 3 aspects:

\begin{itemize}  
\item [1)]
We devised a multi-scale pretext task 
for CL in PolSAR image classification,
which incorporates self-distillation for local-to-global correspondences.

\item [2)]
With the task-tailored Mamba-based block 
and the effective feature fusion module called Cross Mamba, 
ECP-Mamba firstly explores 
the application of SSMs in PolSAR image classification.

\item [3)]
ECP-Mamba manages to balance the performance and efficiency, 
offering cutting-edge performance 
with much lower FLOPs.
\end{itemize}

\begin{figure*}[htb]
    \vspace{-0.1cm}
    \begin{center}
    \includegraphics[width=1.0\textwidth]{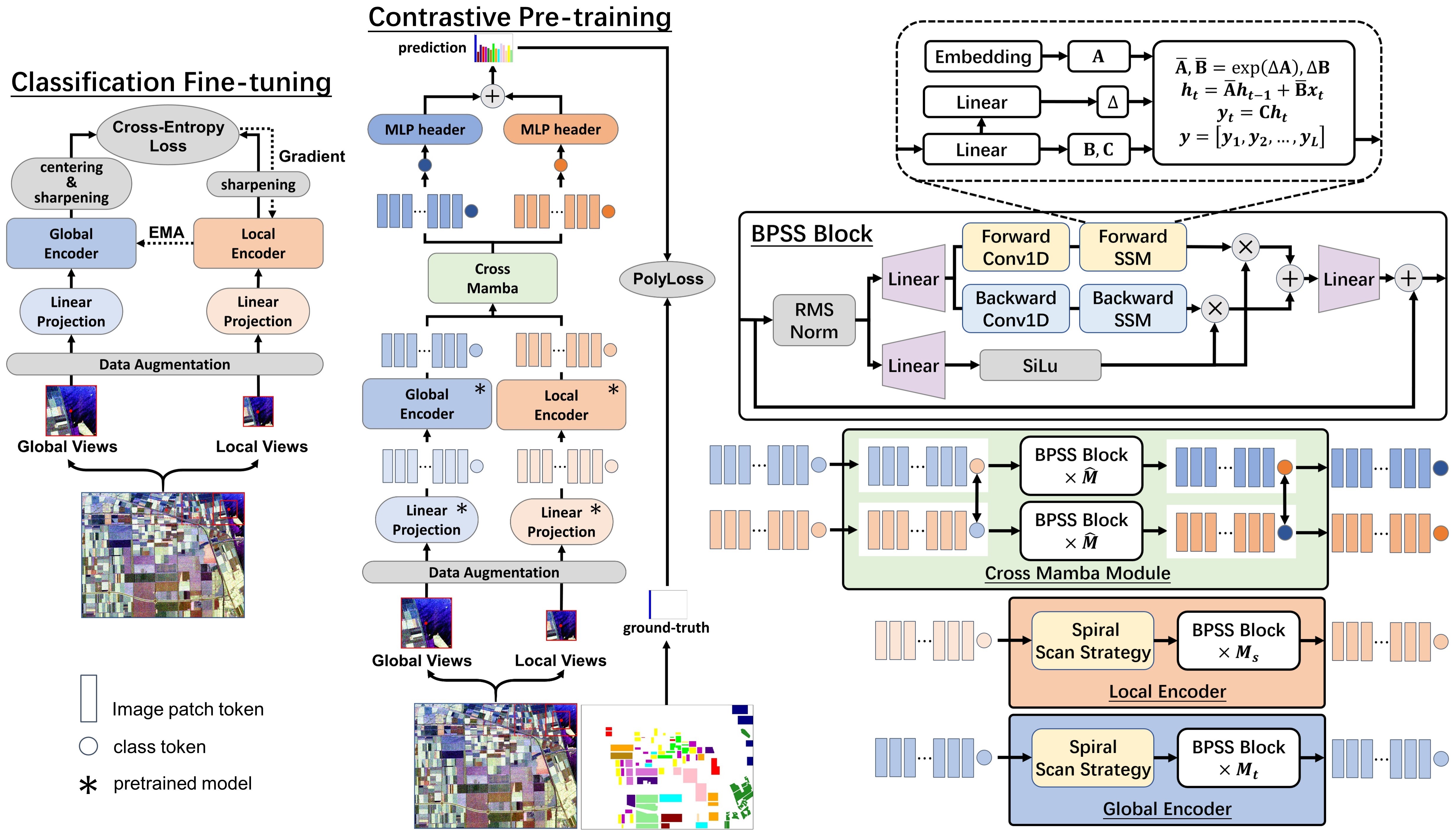}
    \vspace{-0.7cm}
    \caption{
    Illustration of ECP-Mamba, 
    which includes
    Contrastive Pre-training and 
    Classification Fine-tune.
    Contrastive Pre-training is an unsupervised self-distillation stage.
    Constructing local views $x$ and global view $X$
    of a target pixel,
    it feeds augmented $(x, X)$ to 
    the local and global encoders respectively,
    which has same Mamba structure with Spiral Scanning
    but distinct training manners. 
    Their output similarity is quantified by Cross-Entropy Loss.
    Classification Fine-tune inherits two pre-trained encoders,
    and introduces a Cross Mamba encoder
    that crosses class tokens between different branches.
    In this stage,
    ECP-Mamba is trained with limited augmented labeled data,
    and optimized by a Poly Loss function 
    for pixel-wise classification.
    }    
    \label{Figurepipeline}
    \end{center}
    \vspace{-0.5cm}
\end{figure*}

\section{Overview of Our Method}\label{Overview}


In this section, we will first define the notations
and outline the workflow of the proposed method.

When accepting a PolSAR image, 
ECP-Mamba extracts pixel vectors from $\mathbf{T}$
and preprocesses them with complex-valued Z-score normalization.
Subsequently, the pixel vector is reshaped into the format 
$I \in \mathbb{R}^{9 \times H \times W}$,
with $9$ denoting the amount of input channels
and $H$ and $W$ are the height and width of image.
In our study,
we input two different sizes of image patches into two network branches respectively.
The local view $x \in \mathbb{R}^{9 \times k \times k}$ is used for the local branch,
while the global view $X \in \mathbb{R}^{9 \times K \times K}$ is used for the global branch,
in which $K$ and $k$ denote sliding window sizes. 
Therefore, the complete dataset of a PolSAR image can be marked as 
$D=\{(x_1, X_1), (x_2, X_2), \ldots, (x_{N}, X_{N})\}$, 
where $N = H \times W$.
Its labeled sample set can be represented as 
$D'=\{(x_1, X_1; y_1), (x_2, X_2; y_2), \ldots, (x_{N'}, X_{N'}; y_{N'})\}$, 
with $N'$ being the number of annotations and $N' \ll N$.
Here, $y_i \in \{1, 2, \ldots, N_c\}$ signifies the class label 
of the central pixel
of sample pair $(x_i, X_i)$,
where $N_c$ is the number of classes.
Our method aims at assigning a prediction $\bar{y}_i$ to each pixel in $D$ and finally outputting all predictions 
$P = \{\bar{y}_1, \bar{y}_2, \ldots, \bar{y}_N\}$.

As illustrated in Figure \ref{Figurepipeline}, 
ECP-Mamba is composed of two processes: 
\textbf{Contrastive Pre-training} and \textbf{Classification Fine-tuning}.
Our design integrates 2 Mamba encoders with varying feature scales
across 2 branches (i.e., local and global branches) 
to achieve a balance in computational costs.
Contrastive pre-training serves as an unsupervised self-distillation phase. 
By constructing local and global views $(x, X)$ of a pixel as positive pairs,
it inputs random augmentations of $(x, X)$ into the local and global encoders,
which possess the network structure but distinct parameters. 
The global branch employs centering normalization 
by calculating means across the batch.
The similarity between output features is quantified 
using Cross-Entropy Loss.
Only the local branch is gradient-updated,
while global parameters are optimized with 
an exponential moving average (EMA) of local parameters.
Classification Fine-tune inherits the pre-trained local and global encoders
and introduces the Cross Mamba module.
In this stage,
ECP-Mamba is trained with only a limited set of labeled data
with data augmentation technologies.
Via exchanging class tokens between 2 branches, 
Cross Mamba effectively fuses multi-scale features. 
Optimized with a Poly Loss function designed for classification,
the model ultimately generates pixel-wise predictions.

We will first introduce the Contrastive Pre-training stage in Section~\ref{Contrastive_Pretraining}, 
and then delineate the Classification Fine-tuning stage in Section~\ref{Classification}.
Finally, the summary and pseudo-code of ECP-Mamba will be presented in Section~\ref{Summary}.

\section{Contrastive Pre-training}
\label{Contrastive_Pretraining}

Contrastive learning is fundamentally about extracting invariant features from large-scale unlabeled datasets. 
To achieve this,
a multi-scale predictive pretext task 
is crafted to uncover the consistent representations necessary 
for PolSAR images in Section \ref{Design}.
Then the State Space Model theories,
Mamba-based network architecture, 
contrastive loss function and optimization
will be described 
in Section \ref{SSM_Preliminaries} to Section \ref{ContrastiveLoss} respectively.

\subsection{Supervision Signal Design}
\label{Design}

Given the textual consistency of PolSAR data, 
there is a strong similarity between multi-scale features of 
the same pixel.
Furthermore, different scales of pixel patches 
possess unique yet vital
geocoding and scattering information,
rendering them an excellent subject for 
supervision signal research.
%
Thereinto,
we developed
a multi-scale predictive pretext task
for PolSAR image classification
under the guidance of \cite{DINO}. 
Our task seeks to learn polarimetric representations
by transitioning learned features from a localized perspective to a comprehensive viewpoint,
thereby implicitly executing CL without negative sample pairs.

More specifically,
as shown in Figure~\ref{Figurepipeline},
the pre-training model includes 
two learnable network branches: 
the global branch and local branch.
The pretext task is designed to leverage the massive unlabeled pixels within a PolSAR image for pre-training purposes.
Given an input sample pair $(x, X)$ from $D$,
data augmentation techniques 
(detailed in Section \ref{pre_and_aug})
are first applied to generate two augmented views
$\tilde{x}$ and $\tilde{X}$. 
Then $\tilde{x}$ and $\tilde{X}$ are transformed 
into sequences of tokens $v$ and $V$ by 
linear projection layers $f_{\theta_s}$ and $f_{\theta_t}$.
Subsequently,
$v$ and $V$ are fed to 
$g_{\theta_s}$ and $g_{\theta_t}$
to extract the representations of sequences,
and output tokens $w$ and $W$.
The global encoder ($g_{\theta_t}$) and local encoder ($g_{\theta_s}$) are structurally identical, 
each possessing distinct parameter sets denoted as $\theta_t$ and $\theta_s$.
It is worth noting that the last token of $w$ and $W$
(i.e., the tail class token in Section \ref{Spiral_Scan}) 
is considered as the probability distributions of classes
and extracted as $p$ and $P$, 
which are then normalized as follows:
\begin{equation}
    \tilde{p} = \operatorname{softmax}(p / \tau_s), \
    \tilde{P} = \operatorname{softmax}((P-C) / \tau_t).
\end{equation}
Here
$\tau_s, \tau_t > 0$ are the temperatures
to sharpen the output distribution of
softmax normalization.
Only the global branch includes the centering normalization operation,
with its center \(C\) computed across the batch and updated using EMA:
\begin{equation}
    C \leftarrow m C + (1-m) \frac{1}{B} 
    \sum_{i=1}^{B} f_{\theta_{t}}(\tilde{X}^{(i)}){,}
\end{equation}
in which $m>0$ denotes the momentum parameter,
$B$ denotes pre-training batch size
and $\tilde{X}^{(i)}$ denotes the $i$-th data within 
a batch of $\tilde{X}$.

\subsection{State Space Model}
\label{SSM_Preliminaries}


SSMs are mathematical frameworks used to describe the evolution of continuous linear systems,
which can be characterized 
by the below state equations:
\begin{equation}
\label{eq:SSM}
\begin{aligned}
    h'(t) &= \mathbf{A} h(t) + \mathbf{B} x(t) {,} \\
    y(t) &= \mathbf{C} h(t) {,}
\end{aligned}
\end{equation}
in which 
$h(t) \in \mathbb{R}^{N}, x(t) \in \mathbb{R}^{L}$
and $y(t) \in \mathbb{R}^{L}$ represent 
hidden states, input signal and response signal respectively.
The time derivative of $h(t)$ is marked as $h'(t)$, 
where $N$ represents the number of dimensions in the latent state space.
$L$ denotes the sequence length.
matrix $\mathbf{A} \in \mathbb{R}^{N \times N}$
serves for the state transitions.
$\mathbf{B} \in \mathbb{R}^{N \times L}$
and $\mathbf{C} \in \mathbb{R}^{L \times N}$
act as the projection matrices.
To tailor the continuous-time SSM for 
discrete deep learning, 
a time scale parameter $\Delta$ is introduced in S4~\cite{gu2021efficiently}
to facilitate the discretization of matrices 
$\mathbf{A}$ and $\mathbf{B}$
into 
$\bar{\mathbf{A}}$ and $\bar{\mathbf{B}}$
using the zero-order hold method,
which is:
\begin{equation}
\begin{aligned}
\bar{\mathbf{A}} &= \exp(\Delta \mathbf{A}) {,} \\
\bar{\mathbf{B}} &= 
(\Delta \mathbf{A})^{-1}(\bar{\mathbf{A}} - \mathbf{I}) \cdot \Delta \mathbf{B} \approx \Delta \mathbf{B} {.}
\end{aligned}
\end{equation}
Then Equation~\ref{eq:SSM} can be discretized as:
\begin{equation}
\begin{aligned}
h_t &= \bar{\mathbf{A}} h_{t-1} + \bar{\mathbf{B}} x_t {,} \\
y_t &= \mathbf{C} h_t {.}
\end{aligned}
\end{equation}
The calculation of 
$\mathbf{y} = \operatorname{SSM}(\mathbf{x}; \overline{\mathbf{A}}, \overline{\mathbf{B}}, \mathbf{C})$ 
can be further represented in a convolutional form as
\begin{equation}
\begin{aligned}
\bar{\mathbf{K}} & =
\left(\mathbf{C} \bar{\mathbf{B}}, 
\mathbf{C} \bar{\mathbf{A}} \bar{\mathbf{B}},
\cdots, 
\mathbf{C} \bar{\mathbf{A}}^{L-1} \bar{\mathbf{B}}\right){,} \\
\mathbf{y} & = \mathbf{x} * \bar{\mathbf{K}}{,}
\end{aligned}
\end{equation}
in which $\bar{\mathbf{K}}$ is structured as the convolutional kernel, 
and $*$ is the convolution operation.
%
To enhance the model's dynamic adjustment capabilities 
based on contexts,
Mamba~\cite{gu2023mamba} designs the selective mechanism 
in $\mathbf{B}, \mathbf{C}$ and $\Delta$, expressed as:
\begin{equation}
\begin{aligned}
\mathbf{B} &= s_{B}(\mathbf{x}){,} \\
\mathbf{C} &= s_{C}(\mathbf{x}){,} \\
\Delta &= \tau_{\Delta} \left(Parameter + s_{\Delta}(\mathbf{x})\right){,}
\end{aligned}
\end{equation}
where both $s_{B}(\cdot)$ and $s_{C}(\cdot)$ are linear layers projecting input to dimension $N$;
$s_{\Delta}(\cdot)$ first projects the input to 1, 
and then broadcasts it to dimension $D$;
$\tau_{\Delta}(\cdot)$ is softplus function~\cite{xu2024visualmambasurveynew}.
Here $\mathbf{x}, \mathbf{y}, \Delta \in \mathbb{R}^{B \times L \times D}$ and
$\mathbf{B}, \mathbf{C} \in \mathbb{R}^{B \times L \times N}$,
in which 
$\mathbf{x}$ and $\mathbf{y}$ denote input and output;
$B$ and $D$ denote the batch size and the number of channels respectively.

\begin{figure*}[htbp]
\vspace{-0.25cm}
    \begin{center}
    \includegraphics[width=0.95\textwidth]{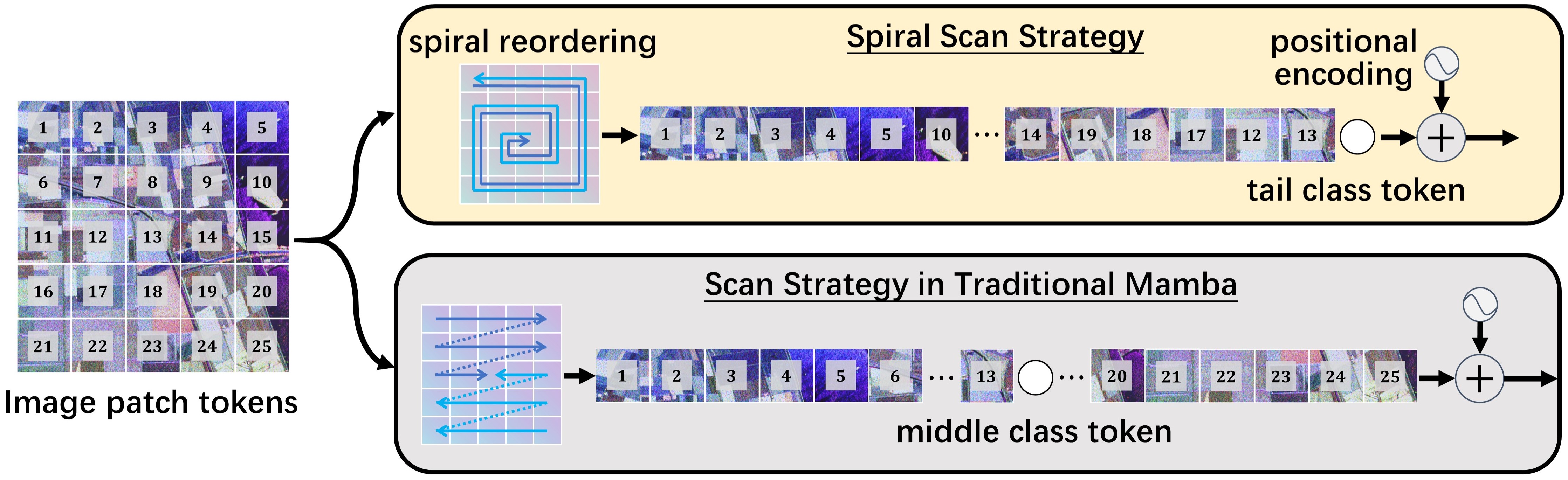}
    \vspace{-0.30cm}
    \caption{
    Illustration of 
    the scan strategy in traditional Mamba~\cite{gu2023mamba}
    and the proposed spiral scan strategy.
    Traditional Mamba uses a sequential scan with raster row order and a middle class token.
    In contrast, our spiral scan strategy introduces the spiral reordering process
    along with a tail class token.
    }
    \label{Figurespiral_scan}
    \end{center}
    \vspace{-0.70cm}
\end{figure*}

\subsection{Mamba-based Network Architecture}

In this part, 
we introduce the devised Mamba-based networks named 
Multi-scale Efficient Mamba
to effectively capture long-range polarimetric relationships
in PolSAR imagery.

Receiving an augmented patch
$\tilde{x} \in \mathbb{R}^{B \times 9 \times k \times k}$ 
as input,
the local branch first reshapes 
the image tensor
into semantic feature sequences
$v \in \mathbb{R}^{B \times L \times D}$
via the following operations:
\begin{equation}
\begin{aligned}
v &= f_{\theta_s}(\tilde{x}) \\
&= \operatorname{Transpose}(\operatorname{Flatten}(\operatorname{Conv2D}(\tilde{x}))){,} \\
\end{aligned}
\end{equation}
where 
$\operatorname{Flatten}(\cdot)$ denotes unfolding the last two dimensions into a single dimension,
and $\operatorname{Transpose}(\cdot)$ denotes swapping the last two dimensions of the tensor.
The kernel size of the 2D convolutional layer 
$\operatorname{Conv2D}(\cdot)$
is defined as $\hat{k}$,
representing the scale of feature characterization.
Similar operations are conducted
on $\tilde{X} \in \mathbb{R}^{B \times 9 \times K \times K}$ 
to obtain $V \in \mathbb{R}^{B \times L \times D}$, 
while using a convolutional kernel size $\hat{K}$.

\subsubsection{Spiral Scan Strategy}
\label{Spiral_Scan} 

Traditional Mamba is typically utilized for causal modeling of sequences,
which struggles to represent spatial and contextual relationships
in imagery scenarios.
%
When doing per-pixel classification on PolSAR images,
researchers usually extract patches from a PolSAR image with a window box, 
with each patch's annotation matching the label of its center.
This emphasizes that features closer to the central pixel
are more causally relevant to the annotation.
Motivated by this insight, 
we propose the spiral scan strategy,
a novel scanning mechanism for Mamba that 
leverages the central pixel's informational significance.


As shown in Figure~\ref{Figurespiral_scan},
when sequences of image tokens are obtained as input, 
$v$ is first reordered by function $\Phi_{\text{spiral}}(\cdot)$.
It executes a spiral traversal
across the second dimension (i.e., $L$ dimension) of 3D tensor,
which starts from the top-left corner, moving rightwards, then downwards and leftwards,
cyclically repeating this pattern
until each element is tracked only once.
Finally,
$\Phi_{\text{spiral}}(v)$ returns 
the rearranged sequences of $v$.
Then a tail class token is appended next to 
the final token,
emphasizing the feature learning
with a stronger causal semantic relationship 
to the annotation.
Afterwards,
positional encoding is added to better handle
the relative spatial information.
When getting $v$,
the whole procedure of spiral scan strategy 
($\operatorname{SpiralScan}$)
can be summarized as follows:
\begin{equation}
\begin{aligned}
v_s &= \operatorname{SpiralScan}(v)\\
&= \operatorname{Concat}(\Phi_{\text{spiral}}(v), c_t) + pos {,}
\end{aligned}
\end{equation}
where $v_s \in \mathbb{R}^{B \times (L+1) \times D}$,
$\operatorname{Concat}(\cdot)$ is concatenation on the length dimension,
$c_t$ is the learnable tail class token,
and $pos$ is defined as sinusoidal positional embedding \cite{ho2020denoising}.

The proposed strategy systematically traverses the image patch 
while ensuring the spatial and semantic continuity of pixels, 
thereby maximizing the performance of SSM mechanism
in PolSAR image classification.

\subsubsection{Bidirectional Polarimetric Selective Scanning Block}
\label{Mamba_Block}

It is essential to consider the heterogeneity of PolSAR data 
in polarimetric feature learning. 
Traditional SSMs tend to apply uniform treatment to
each pixel using static convolutional kernels.
%
Based on Vim~\cite{zhu2024vision},
we introduce bidirectional selective scanning to our field,
called Bidirectional Polarimetric Selective Scanning (BPSS) block.
The BPSS block allows each pixel to interact with its surroundings based on the input. 
This enhances discriminative representations through sequential SSM interactions and 
improves the model's ability to extract polarimetric features from various PolSAR targets.

As shown in Figure~\ref{Figurepipeline}, 
obtaining $v_s$ as input,
the workflow of BPSS block is shown as follows:
\begin{equation}
\begin{aligned}
s & =\operatorname{Linear}_1(v_s){,} \\
\mathcal{F}_{f} & =\operatorname{SSM}_f\left(\operatorname{SiLu}\left(\operatorname{Conv1D}_f\left( s \right)\right)\right){,} \\
\mathcal{F}_{b} & =\operatorname{Flip}\left(\operatorname{SSM}_b\left(\operatorname{SiLu}\left(\operatorname{Conv1D}_b\left(\operatorname{Flip}\left(s
\right)\right)\right)\right)\right){,} \\
\mathcal{G} & =\operatorname{SiLu}(\operatorname{Linear}_2(\operatorname{Norm}(v_s))){,}
\end{aligned}
\end{equation}
where $\operatorname{Linear}(\cdot)$ denotes linear layers,
and $\operatorname{Norm}(\cdot)$ is initialized as RMSNorm~\cite{gu2023mamba}.
In this context, 
the selection mechanism is applied in both forward and reverse directions 
using dimensional reversal operation $\operatorname{Flip}(\cdot)$ along the length axis. 
Both directions undergo 1D convolution layers denoted as $\operatorname{Conv1D}(\cdot)$.
$\operatorname{SSM}_f$ and $\operatorname{SSM}_b$ represent
state space models 
with same structure yet distinct parameters 
$\overline{\mathbf{A}}, \overline{\mathbf{B}}, \mathbf{C}$
in Section~\ref{SSM_Preliminaries},
which is the core of selection mechanism.
Finally, via employing a residual connection to 
maintain the existing polarimetric correlation, 
the output of BPSS block can be obtained as
\begin{equation}\label{ResidualConnection}
v_s' =\operatorname{Linear}_3\left(\mathcal{F}_{f} \odot \mathcal{G}+\mathcal{F}_{b} \odot \mathcal{G}\right) + v_s {,}
\end{equation}
where $v_s' \in \mathbb{R}^{B \times (L+1) \times D}$ and
$\odot$ represents the element-wise multiplication.
It should be noted that 
there are 
$M_s$ and $M_t$ BPSS blocks in 
the local encoder $g_{\theta_s}$ and 
global encoder $g_{\theta_t}$ respectively, 
operating on different scales of features
and producing 
$w, W \in \mathbb{R}^{B \times (L+1) \times D}$.

\subsection{Contrastive Loss Function and Optimization}
\label{ContrastiveLoss}

As elaborated in Section \ref{Design},
in order to enhance the local-to-global correspondences, 
we aim to align the distributions 
between $\tilde{p}$ and $\tilde{P}$
by minimizing the Cross-Entropy Loss 
as follows:
\begin{equation}
\hspace{-0.35cm}
\label{Pretraining_Loss}
\begin{aligned} 
\mathcal{L}_1 = 
- \frac{1}{D} \sum_{i=1}^{D} \tilde{P}^{(i)} \log(\tilde{p}^{(i)}){,}  
\end{aligned}
\end{equation}
where $\tilde{P}^{(i)}$ and $\tilde{p}^{(i)}$ are the 
$i$-th elements of $\tilde{p}$ and $\tilde{P}$ over $D$ channel dimensions
respectively.

Contrastive Pre-training process aims to pre-train two encoders
($g_{\theta_s}$, $g_{\theta_t}$)
with their projectors 
($f_{\theta_s}$, $f_{\theta_t}$)
by optimizing the parameters 
$\theta_s$ and $\theta_t$ 
for the subsequent Classification process.
As depicted in Figure~\ref{Figurepipeline}, 
gradient back propagation is executed only on $\theta_s$ during the pre-training, 
while $\theta_t$ is updated as the EMA of $\theta_s$,
that is controlled by a momentum parameter $\lambda$.
The $\lambda$ follows a cosine schedule increasing from $0.9995$ to $1$ during training according to \cite{DINO}.
In summary, the optimization dynamics of Contrastive Pre-training can be formulated as follows:
\begin{equation}
\begin{aligned}
    \theta_{s} &\leftarrow \operatorname{AdamW}\left(\theta_{s}, \nabla_{\theta_{s}}\mathcal{L}_1, \eta_{1}\right), \\
    \theta_{t} &\leftarrow \lambda \theta_{t} + (1-\lambda) \theta_{s},
\end{aligned}
\end{equation}
where $\operatorname{AdamW}$ \cite{AdamW} is utilized as the optimizer 
to accelerate convergence and mitigate overfitting,
and $\eta_{1}$ represents the pre-training learning rate.

\section{Classification Fine-tuning}
\label{Classification}

In Classification Fine-tuning process,
both Mamba-based encoders
and their linear projections 
will be fine-tuned for 
the downstream classification task
with only a limited set of human annotations,
whose parameters have been pre-trained 
from $\theta_s, \theta_t$ to $\Theta_s, \Theta_t$.
In this section,
we will first delineate 
the proposed Cross Mamba module in Section \ref{CrossMamba}, 
and then introduce the classification framework 
including loss function and optimization 
in Section \ref{Classification_Framework}.
Finally, data preprocessing and augmentation are detailed in Section \ref{pre_and_aug}.

\subsection{Cross Mamba Module}
\label{CrossMamba}

Although each pre-trained encoder can extract features at a specific scale,
there is a lack of interactive integration between these features, 
which is critical for enhancing the complementarity between multi-scale features 
and improving classification performance.
Indeed, cross-branch token interactions enhance feature representation across scales,
offering a more comprehensive multi-scale perspective for SSMs.
Meanwhile, class tokens \(w\) and \(W\)
share identical dimensions and are spatially arranged in the same spiral pattern. 
Positioned terminally within the sequence, 
their class tokens represent the final temporal state output of their respective SSMs, 
having aggregated information from all patch tokens at their scale.

\begin{figure}[htbp]
\vspace{-0.10cm}
    \begin{center}
    \includegraphics[width=0.475\textwidth]{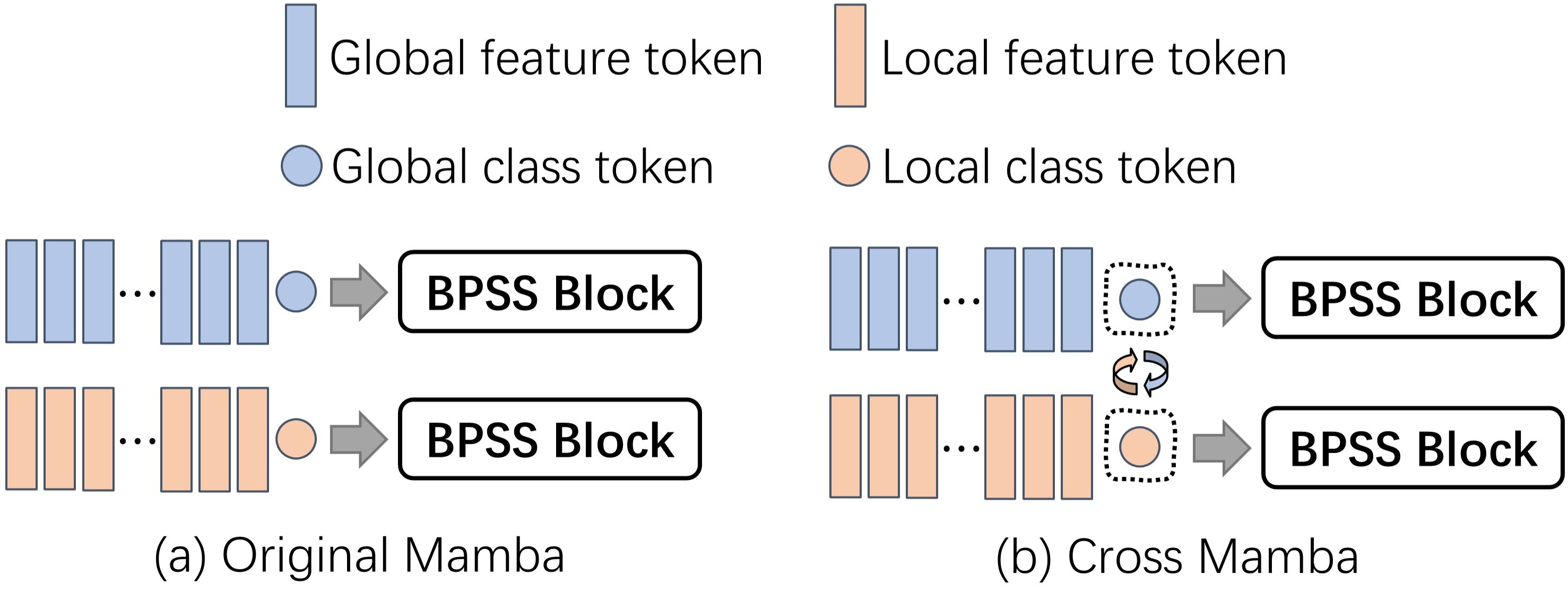}
    \vspace{-0.25cm}
    \caption{
    (a) Traditional Mamba without feature fusion.
    (b) The proposed Cross Mamba module exchanges class tokens between branches to introduce multi-scale feature interaction.
    }
    \label{Fig:CrossMamba}
    \end{center}
    \vspace{-0.25cm}
\end{figure}

Building on these, 
we present the multi-scale feature fusion module named Cross Mamba.
As shown in Figure~\ref{Fig:CrossMamba}, 
considering spatial correlations and contextual cues across different scales, 
this module employs a straightforward yet effective approach. 
Before feeding sequences to $\hat{M}$ layers of BPSS blocks, 
Cross Mamba exchanges the last token of input sequences 
(i.e., the tail class tokens) 
between local and global branches.
By integrating cross-branch features,
it overcomes limitations of single-dimension 
spatial state propagation, 
enabling more comprehensive patch representation.





\subsection{Classification Framework}
\label{Classification_Framework}

In Classification Fine-tuning,
we restrict the training to a curated set of labeled data,
where each instance \(y\) corresponds to a local view \(x\) and a global view \(X\) from \(D'\). 
By applying data augmentation techniques in Section \ref{pre_and_aug},
we generate augmented view pairs \(\tilde{x}\) and \(\tilde{X}\).
Parallel to the feature extractions of
Section \ref{Design},
the feature sequences are first derived as follows:
\begin{equation}
\begin{aligned}
w &= g_{\Theta_s}(f_{\Theta_s}(\tilde{x})) {,} \\ 
W &= g_{\Theta_t}(f_{\Theta_t}(\tilde{X})) {.}
\end{aligned}
\end{equation}
They are then fed to Cross Mamba module
that 
delivers multi-scale integrated features \(w'\) and \(W'\).
This process is defined as 
\(w', W' = \hat{g}_{\Theta_s, \Theta_t}(w, W)\),
whose details are provided in Section \ref{CrossMamba}.
Only the final tokens from \(w'\) and \(W'\),
denoted as \(c_t\) and \(C_t\), 
are used to predict category probability distributions. 
Two multilayer perceptrons (MLPs),
$h_{\Theta_s}$ and $h_{\Theta_t}$,
serve as classification headers,
with the network generating predictions $\hat{y}$ and $\hat{Y}$ for each branch.
Ultimately, the output integrates and averages 
the predictive information from both scale branches as 
$\bar{y} = (\hat{Y}+\hat{y}) / 2$, here $\bar{y} \in \mathbb{R}^{B \times N_c}$.
%
The whole classification process can be simplified as
$\bar{y} = F_{\Theta_s, \Theta_t}(x, X)$,
where $F_{\Theta_s, \Theta_t}$ denotes the classifier.

As the polynomially-weighted variant of the Cross-Entropy loss, 
PolyLoss~\cite{leng2022polyloss} is particularly designed for classification tasks. 
We implement this advanced and adaptable loss function within the fine-tuning.
This monotonically decreasing function can be expressed as:
\begin{equation}
\begin{aligned}
\mathcal{L}_2 
&= \frac{1}{N_c} \sum_{i=1}^{N_c} 
\Big[
(1 + \epsilon_1)(1 - \bar{y}^{(i)}) +
\sum_{j=1}^{\infty} \frac{1}{j+1} \left( 1-\bar{y}^{(i)} \right)^{j+1}
\Big] \\
&= \frac{1}{N_c} \sum_{i=1}^{N_c} 
\Big[
- y^{(i)} \log(\bar{y}^{(i)})
+ \epsilon(1 - \bar{y}^{(i)})
\Big] {.}
\end{aligned}
\end{equation}
It reduces to the Cross-Entropy loss
when the polynomial coefficient $\epsilon$ is 0.
$\bar{y}^{(i)}$ and $y^{(i)}$ correspond to the $i$-th components of the predicted values $\hat{y}$ and true labels $y$, respectively.
The initial label is one-hot encoded into 
a $N_c$-dimensional vector $y$,
whose component corresponding to the class is set to $1$,
while all others are $0$.
The predicted class for $\bar{y}$ is determined by the index $i$ of its maximum component $\bar{y}^{(i)}$.
Finally, we input $D$ to classifier $F_{\Theta_s, \Theta_t}$ 
and output pixel-level classification result $O$.
The model parameters are optimized using AdamW, as shown below:
\begin{equation}
    \Theta_{s}, \Theta_{t} \leftarrow \operatorname{AdamW}\left(\Theta_{s}, \Theta_{t}, \nabla_{\Theta_{s}, \Theta_{t}}\mathcal{L}_2, \eta_{2}\right) {,}
\end{equation}
where $\eta_{2}$ represents the fine-tuning learning rate.

\subsection{Data Preprocessing and Augmentation}
\label{pre_and_aug}

In this part,
we will introduce the 
data preprocessing and data augmentation technologies 
for ECP-Mamba.

Considering the integrity of polarimetric information, 
the upper triangular elements of $\mathbf{T}$ matrix in Equation \ref{T},
i.e., ${\left[{T}_{11}\ {T}_{22}\ {T}_{33}\ {T}_{12}\ {T}_{13}\ {T}_{23}\right]}$ 
are used as raw input data in this work.
To facilitate model convergence,
it is imperative to normalize each vector
using complex-valued Z-score standardization.
For channel $\mathcal{T}$ whose $i$-th element is denoted as
$\mathcal{T}^{(i)}$,
its mean value $\mathcal{T}_{avg}$ 
and standard deviation value $\mathcal{T}_{std}$ are calculated as:
\begin{equation}
\begin{aligned}
\mathcal{T}_{avg} & =\frac{\sum_{i=1}^{N} \mathcal{T}^{(i)}}{N} \\
\mathcal{T}_{std} & =\sqrt{\frac{\sum_{i=1}^{N}\left(\mathcal{T}^{(i)}-\mathcal{T}_{avg}\right) {\left(\mathcal{T}^{(i)}-\mathcal{T}_{avg}\right)}^{\ast}}{N}},
\end{aligned}
\end{equation}
in which $N$ is the total number of vectors,
and $^*$ denotes the conjugate operation.
Following the standardization process,
the vectors are restructured into the following format:
\begin{equation}
\begin{aligned}
\big[ \ {T}_{11}\ {T}_{22}\ {T}_{33}\
& \Re({T}_{12})\
\Re({T}_{13})\ 
\Re({T}_{23}) \\
& \Im({T}_{12})\
\Im({T}_{13})\
\Im({T}_{23})\ \big]{,}
\end{aligned}
\end{equation}
where $\Re(\cdot)$ and $\Im(\cdot)$ denote 
the real and imaginary components respectively. 
This configuration finally yields the input tensor 
$I \in \mathbb{R}^{9 \times H \times W}$.

Data augmentation is a critical component in network training, 
significantly influencing the accuracy and robustness of classification models. 
We expand the training dataset by generating 7 distinct positive samples 
through specific augmentation techniques applied to the original image patch, 
which include mirror flipping,
rotations of the original and its flipped version 
by 90, 180, and 270 degrees. 
These operations enrich the dataset and contribute to the model's ability to learn from varied perspectives, thereby enhancing its generalization capabilities.

\section{Summary of Our Method}
\label{Summary}

The workflows for 
Contrastive Pre-training
(see Section~\ref{Contrastive_Pretraining})
and Classification Fine-tuning stage
(see Section~\ref{Classification})
are depicted in Algorithm~\ref{algorithm:Contrastive_Pre-training} and Algorithm~\ref{algorithm:Classification} respectively.

In Contrastive Pre-training process, 
given a PolSAR image as input, 
the pre-training phase is initiated by unsupervised training of the global and local branches
under the guidance of 
the multi-scale predictive pretext task 
(see Section~\ref{Design}),
which employ spiral scan strategy 
(see Section~\ref{Spiral_Scan})
and BPSS blocks
(see Section~\ref{Mamba_Block}).
%
The Classification Fine-tuning process
leverages pre-trained models.
To enhance the interaction of different scales,
the Cross Mamba module
(see Sections~\ref{CrossMamba})
is applied following the implementation of
pre-trained encoders.
%
Finally, 
the classifier is trained to yield the pixel-wise predictions of PolSAR image
(see Section~\ref{Classification_Framework}).
%
We apply data preprocessing and augmentation throughout both processes
to standardize data and expand training data 
(see Section~\ref{pre_and_aug}).

\begin{figure*}[htbp]
        \begin{center}
            \includegraphics[width=0.90\textwidth]{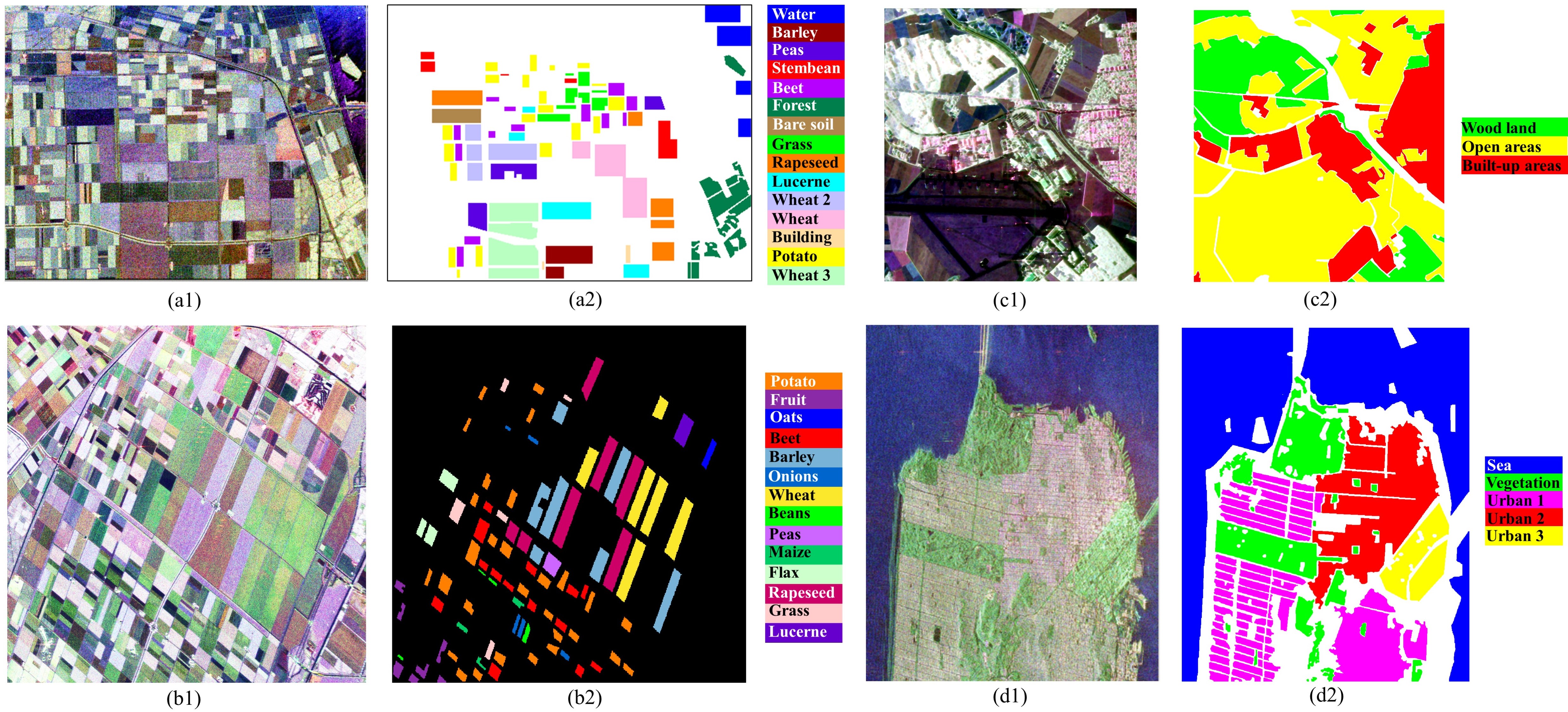}
            \vspace{-0.425cm}
            \caption
            {
            This figure visually represents the Pauli RGB images and their corresponding ground truth label maps for four PolSAR datasets.
            (a1) the Pauli RGB image for the Flevoland 1989 dataset,
            with (a2) its ground truth map and legend. 
            (b1) the Pauli RGB image for the Flevoland 1991 dataset, 
            with (b2) its ground truth map and legend.
            (c1) the Pauli RGB image for the Oberpfaffenhofen dataset, 
            with (c2) its ground truth map and legend.
            (d1) the Pauli RGB image for the San Francisco dataset, 
            with (d2) its ground truth map and legend.
            }
            \vspace{-0.5cm}
            \label{fig:Experimental_images}
        \end{center}
\end{figure*}

\begin{algorithm}[htb]
\caption{\emph{Contrastive Pre-training}}
\label{algorithm:Contrastive_Pre-training}

\SetKwInOut{Input}{Input}\SetKwInOut{Output}{Output}
\Input{dataset $D$ \\
linear projection layers $f_{\theta_s}$, $g_{\theta_t}$ \\
Mamba encoders $g_{\theta_s}$, $g_{\theta_t}$ \\
sharpening temperatures $\tau_s$, $\tau_t$ \\
centering parameters $C$, $m$ \\
momentum parameter $\lambda$ \\
learning rate $\eta_{1}$ for pre-training\\ 
epoch number $\mathcal{I}_{1}$ for pre-training
} 
\Output{pre-trained $f_{\theta_s}, f_{\theta_t}, g_{\theta_s}, g_{\theta_t}$ for Algorithm 2}
\label{alg:2}
\begin{algorithmic}

    \FOR{iteration in $\mathcal{I}_{1}$}
    
    \STATE Randomly select $(x, X)$ from $D$

    \STATE $\tilde{x}, \tilde{X} =
    \operatorname{augment}(x), \operatorname{augment}(X)$

    \STATE $v, V = f_{\theta_s}(\tilde{x}), f_{\theta_t}(\tilde{X})$

    \STATE $w, W = g_{\theta_s}(v), g_{\theta_t}(V)$

    \STATE Extract class token $p, P$ from $w, W$

    \STATE
    $\tilde{p} = \operatorname{softmax}(p / \tau_s), \
    \tilde{P} = \operatorname{softmax}((P-C) / \tau_t)$

    \STATE 
    Compute contrastive loss $\mathcal{L}_1$
    
    \STATE 
    $\theta_{s} \leftarrow \operatorname{AdamW}\left(\theta_{s}, \nabla_{\theta_{s}}\mathcal{L}_1, \eta_{1}\right)$

    \STATE
    $\theta_{t} \leftarrow \lambda \theta_{t} + (1-\lambda) \theta_{s}$
    
    \ENDFOR

    \RETURN
    $f_{\theta_s}, f_{\theta_t}, g_{\theta_s}, g_{\theta_t}$

\end{algorithmic}  
\end{algorithm}
\begin{algorithm}[htb]
\caption{\emph{Classification Fine-tuning}}
\label{algorithm:Classification}

\SetKwInOut{Input}{Input}\SetKwInOut{Output}{Output}
\Input{dataset $D$ \\
labeled dataset $D'$ \\
load     
$f_{\theta_s}, f_{\theta_t}, g_{\theta_s}, g_{\theta_t}$  
as 
$f_{\Theta_s}, f_{\Theta_t}, g_{\Theta_s}, g_{\Theta_t}$ \\
Cross Mamba module $\hat{g}_{\Theta_s, \Theta_t}$ \\
MLP headers $h_{\Theta_s}, h_{\Theta_t}$ \\
learning rate $\eta_{2}$ for fine-tuning \\
epoch number $\mathcal{I}_{2}$ for fine-tuning }

\Output{classifier $F_{\Theta_s, \Theta_t}$ \\
pixel-level classification result $O$}

\label{alg:1}
\begin{algorithmic}
    
    \FOR{iteration in $\mathcal{I}_{2}$}
    
    \STATE Randomly select $(x, X)$ from $D'$

    \STATE $\tilde{x}, \tilde{X} =
    \operatorname{augment}(x), \operatorname{augment}(X)$

    \STATE
    $w, W = g_{\Theta_s}(f_{\Theta_s}(\tilde{x})), 
    g_{\Theta_t}(f_{\Theta_t}(\tilde{X}))$

    \STATE
    $w', W' = \hat{g}_{\Theta_s, \Theta_t}(w, W)$

    \STATE Extract class token $c_t, C_t$ from $w', W'$

    \STATE
    $\hat{y}, \hat{Y} = h_{\Theta_s}(c_t), h_{\Theta_t}(C_t)$

    \STATE $\bar{y} = (\hat{Y}+\hat{y}) / 2$

    \STATE
    Compute classification loss $\mathcal{L}_2$

    \STATE
    $\Theta_s, \Theta_t \leftarrow
    \operatorname{AdamW}
    \left(\Theta_{s}, \Theta_t, 
    \nabla_{\Theta_{s}, \Theta_t}\mathcal{L}_2,
    \eta_{2}\right)$

    \ENDFOR

    \RETURN Input $D$ to classifier $F_{\Theta_s, \Theta_t}$ and output $O$
\end{algorithmic}  
\end{algorithm}

\section{Experiments and Analysis}
\label{Experiments}

In this section, 
the effectiveness and superiority of our method 
are demonstrated through its application to 4 benchmark datasets.
The datasets and experimental configurations are first introduced in 
Section \ref{Parameter Setting}. 
In Section \ref{Ablation Study}, 
ablation studies are performed to 
highlight the contributions of each component of ECP-Mamba.
Then a comparative analysis of 
ECP-Mamba against other methods 
is provided in Section \ref{Comparisons and Results}.
Finally, the impact of varying parameters on classification performance
is discussed in Section \ref{Parameter Analysis}.
The model's accuracy performance  
is evaluated using 
Overall Accuracy (OA), 
Average Accuracy (AA)
and Kappa coefficient (Kappa).
The complexity is evaluated using network parameters and FLOPs.
Figure~\ref{fig:Experimental_images} 
presents the Pauli RGB images and 
ground truth labels with their legends 
for 4 benchmark datasets.
Their main properties are provided in Table~\ref{Dataset}. 

\begin{table}[htp]
\renewcommand\arraystretch{1.5}
\caption{Main Information of the Experimental Datasets}
    \resizebox{1\linewidth}{!}
    {
        \begin{tabular}{c|cccccc}
            \toprule
            \hline
            Name & Sensor & Time & Image band & Dimensions & Class
            \\
            \midrule
            Flevoland 1989 & AIRSAR & 1989 & L-band & 750 × 1024 & 15
            \\
            Flevoland 1991 & AIRSAR & 1991 & L-band & 1020 × 1024 & 14
            \\
            Oberpfaffenhofen & E-SAR & - & L-band & 1200 × 1300 & 3
            \\
            San Francisco & RADARSAT-2 & 2008 & C-band & 1380 × 1800 & 5 
            \\
            \hline
            \bottomrule
        \end{tabular}
     }
\label{Dataset}
\end{table}

\subsection{Parameter Setting}
\label{Parameter Setting}

\begin{table*}
\caption{Metrics on the Flevoland 1989 Dataset for Ablation Study on ECP-Mamba}
\renewcommand\arraystretch{1.2}
\resizebox{1.0\linewidth}{!}
{
    {\begin{threeparttable}
        \begin{tabular}{cccccccccccc}
        \toprule
        \hline        
        \multirow{3}{*}{Model} & 
        \multicolumn{5}{c}{Components} &
        \multirow{3}{*}{OA(\%)} &
        \multirow{3}{*}{AA(\%)} &
        \multirow{3}{*}{Kappa(e-2)} &
        \multirow{3}{*}{Params(M)} &
        \multirow{3}{*}{FLOPs(M)} &
        \\        
        \cmidrule {2-6} &
        \makecell[c]{Spiral \\ Scan} &   
        \makecell[c]{Multi-scale \\ Branch} &
        \makecell[c]{Cross \\ Mamba} &
        \makecell[c]{Contrastive \\ Pre-training} &
        \makecell[c]{Data \\ Augmentation} & 
        \\ 
        \midrule
        ViT &\ding{56} &\ding{56} &\ding{56} &\ding{56} &\ding{56} & 89.94 & 88.34 & 87.50 & 2.20 & 147.05 \\
        Vim &\ding{56} &\ding{56} &\ding{56} &\ding{56} &\ding{56} & 91.75 & 90.52 & 89.84 & 2.17 & 15.01 \\
        Spiral Mamba &\ding{52} &\ding{56} &\ding{56} &\ding{56} &\ding{56} & 97.30 & 96.61 & 96.37 & 2.17 & 15.06 \\
        Multi-scale Spiral Mamba &\ding{52} &\ding{52} &\ding{56} &\ding{56} &\ding{56} & 98.05 & 97.44 & 97.25 & 4.36 & 31.44 \\
        Multi-scale Efficient Mamba &\ding{52} &\ding{52} &\ding{52} &\ding{56} &\ding{56} & 98.58 & 97.62 & 97.45 & 4.36 & 31.44 \\
        \midrule       
        CL + Multi-scale Efficient Mamba &\ding{52} &\ding{52} &\ding{52} &\ding{52} &\ding{56} & 99.65 & 99.60 & 99.58 & 4.36 & - \\
        ECP-Mamba &\ding{52} &\ding{52} &\ding{52} &\ding{52} &\ding{52} & \textbf{99.70} & \textbf{99.64} & \textbf{99.62} & 4.36 & - \\
        \hline
        \bottomrule
        \end{tabular}
    
        \begin{tablenotes}
            \footnotesize 
            \item[*] 
            The best accuracy results are shown in bold.
            The `-' in the table means that FLOPs does not apply to the self-supervised method.
        \end{tablenotes}
    
    \end{threeparttable}
    }
}
\label{Ablation_Study_table}
\end{table*}

The dimensions of local and global views, 
denoted by \(k\) and \(K\),
are 16 and 32,
with their central pixel coordinates at (8, 8) and (16, 16) respectively. 
To manage pixels adjacent to image edges, 
zero-padding is implemented.
The kernel sizes \(\hat{k}\) and \(\hat{K}\) are 1 and 2 respectively, 
resulting in \(L = 256\). 
The hidden dimension \(D\) is 192. 
The number of BPSS layers in 
\(g_{\theta_s}\), \(g_{\theta_t}\) and \(\hat{g}_{\Theta_s, \Theta_t}\) 
(i.e., $M_s$, $M_t$ and $\hat{M}$)
is uniformly set to 1. 
We will analyze 
the aforementioned hyper-parameter settings 
in Section \ref{parameter_analysis1}.

In Contrastive Pre-training,
we pre-train the model for 100 epochs 
($\mathcal{I}_{1}$),
whose batch size $B$ is set as 128.
The network is optimized 
with a learning rate $\eta_{1}$ of 0.0005. 
The sharpening temperatures $\tau_s$ and $\tau_t$ are set to 0.1 and 0.04 respectively. 
The centering momentum parameter $m$ is initialized at 0.9,
and the centering parameter $C$ is initialized to 0. 
The momentum parameter $\lambda$ is 0.996.
In Classification Fine-tuning, 
a small subset of labeled pixels
is randomly chosen for training.
Following the sampling rate (SR) studies of \cite{kuang2024polarimetry}, 
a SR of 0.05\% is utilized for 
the San Francisco dataset,
while other datasets are applied with a SR of 0.2\%.
The model is fine-tuned over 100 epochs
($\mathcal{I}_{2}$),
with a batch size of 128 and 
a learning rate $\eta_{2}$ of 0.001.
The learning rate and momentum parameters undergo decay following 
a cosine annealing strategy~\cite{cosineAnnealingLR_in_DNNs}, 
which is applied across both processes.
All experimental procedures are conducted within the PyTorch framework
on a workstation with 
one NVIDIA GeForce RTX 4090 GPU
possessing 24-GB memory.

\subsection{Ablation Study}\label{Ablation Study}

To examine the effectiveness of each component of
our proposed ECP-Mamba, 
we conduct 7 groups of experiments on the
Flevoland 1989 dataset as follows:

\begin{itemize}

\item[$\bullet$] \textbf{ViT}: 
A light-scaled Vision Transformer used as the baseline model for architecture comparison,
whose training settings are the same as local branch.

\item[$\bullet$] \textbf{Vim}: 
A basic Vision Mamba without enhancements for PolSAR image classification tasks, 
with parameter settings identical to the local branch.

\item[$\bullet$] \textbf{Spiral Mamba}: 
A Mamba-based model using spiral scan strategy 
to enhance feature extraction capabilities.
The parameter settings are the same as local branch.

\item[$\bullet$] \textbf{Multi-scale Spiral Mamba}: 
An extension of Spiral Mamba, 
enhanced by multi-scale feature extraction branches to 
capture local and global features respectively.

\item[$\bullet$] \textbf{Multi-scale Efficient Mamba}: 
A variant of Multi-scale Spiral Mamba,
which introduces Cross Mamba module  
to considering cross-scale feature interactions.

\item[$\bullet$] \textbf{CL + Multi-scale Efficient Mamba}: 
A self-supervised version of Multi-scale Efficient Mamba,
using a multi-scale predictive pretext task 
for local-to-global correspondence pre-training.

\item[$\bullet$] \textbf{ECP-Mamba}: 
Our proposed method,
which includes Multi-scale Efficient Mamba, contrastive learning and data augmentation technologies.

\end{itemize}

Table~\ref{Ablation_Study_table} presents the numerical results of experiments. 
The following key findings can be highlighted from it:

\begin{enumerate}
\renewcommand{\theenumi}{\alph{enumi}}
\renewcommand{\labelenumi}{\theenumi)}

\item
Compared to the ViT baseline, the Spiral Mamba shows improvements of 7.36\%, 8.27\% and 8.87e-2 
in OA, AA and Kappa values,
while utilizing 98.64\% of network parameters 
and 10.24\% of computational complexity.
In comparison with Vim, 
Spiral Mamba gains
improvements of 5.55\%, 6.09\% and 6.53e-2 in OA, AA and Kappa values
with only 0.33\% FLOPs increase.
These improvements underscore the effectiveness of spiral scan strategy in Mamba structure, 
enhancing its ability to leverage the significance of central pixel.

\item
Combining local and global features through multi-scale branches,
the Multi-scale Spiral Mamba
improves performance by 0.93\% (OA), 3.85\% (AA) and 4.12e-2 (Kappa) over the Spiral Mamba.
The introduction of Cross Mamba
results in additional improvements of 
0.43\% (OA), 0.18\% (AA) and 0.19e-2 (Kappa)
while maintaining parameter and computational efficiency
compared to the Multi-scale Spiral Mamba,
which demonstrates the effectiveness of Cross Mamba 
in refining features across different scales.

\item
CL scheme further enhances the performance of Multi-scale Efficient Mamba.
By integrating CL, the model shows improvements of 1.07\% (OA), 2.02\% (AA) and 2.13e-2 (Kappa) 
over the Multi-scale Efficient Mamba, 
highlighting the utility of self-supervision in scenarios with scarce annotations. 
In conclusion,
ECP-Mamba achieves 7.95\% (OA), 9.12\% (AA) and 9.78e-2 (Kappa) improvements over Vim, 
affirming the benefits of the proposed components in attaining superior performance with efficient resource utilization.
\end{enumerate}

\begin{figure}[htp]
        \begin{center}
        \includegraphics[width=0.45\textwidth]{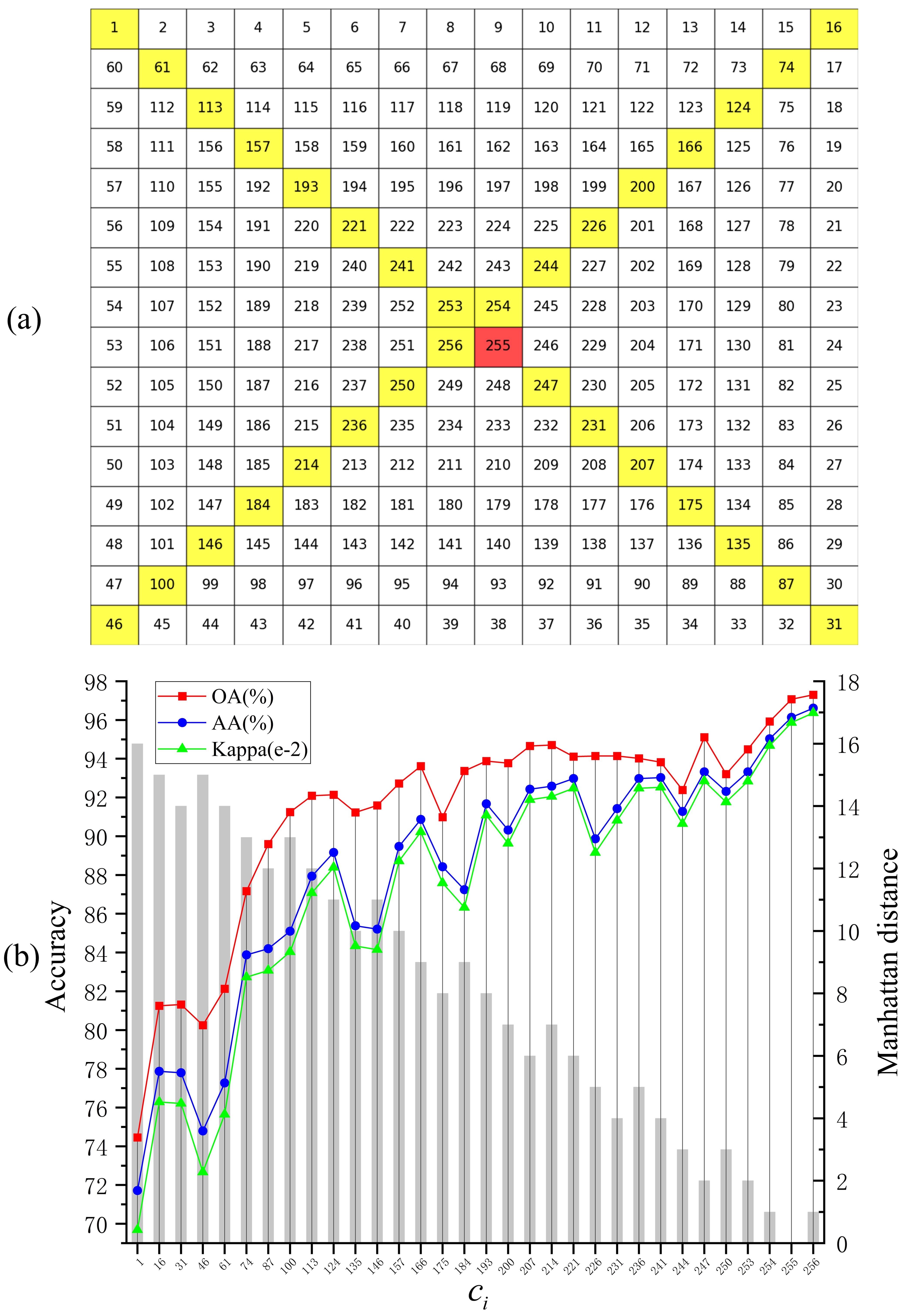}
        \vspace{-0.37cm}
        \caption{
        (a) Visualization of the proposed 
        spiral scan strategy 
        within a \(16 \times 16\) patch. 
        Yellow marks indicate the evaluated positions, 
        and the central pixel's token index is marked in red.
        (b) OA, AA and Kappa curves 
        of Spiral Mamba
        vary with \(c_i\)
        on the Flevoland 1989 dataset,
        alongside the spatial Manhattan distances 
        of corresponding pixels from the central pixel.
        }
        \label{fig:visualization}
        \end{center}   
\end{figure}

To evaluate the superiority of 
our proposed spiral scan strategy, 
we conduct experiments 
by positioning the class token $c_t$
at various locations 
within the sequence $\Phi_{\text{spiral}}(v)$
in Spiral Mamba.
More specifically, insertion of $c_t$ 
after the $i$-th token of $\Phi_{\text{spiral}}(v)$
is defined as position $c_i$, 
where the tail class token position 
in Section~\ref{Spiral_Scan} 
corresponds to $c_t=c_{L}$.
For experimental simplification,
we restrict our selection to diagonal pixel indices
in the original spatial arrangement of patches. 
As visualized in Figure~\ref{fig:visualization}(a),
within a $16 \times 16$ patch, 
the evaluated positions are highlighted in yellow,
while the central pixel's token index is marked in red.
Figure~\ref{fig:visualization}(b) presents 
the OA, AA and Kappa curves
of Spiral Mamba
for varied $c_i$, 
alongside the spatial Manhattan distances of
corresponding pixels from the central pixel. 
Based on Figure~\ref{fig:visualization}, 
we draw the following conclusions:

\begin{enumerate}
\renewcommand{\theenumi}{\alph{enumi}}
\renewcommand{\labelenumi}{\theenumi)}

\item 
The proposed spiral scanning order 
is better aligned with the computation process 
of state space models, 
as it effectively captures local textures 
through short-range state interactions.
Distant tokens like \( c_{1} \)
are less effective than \( c_{256} \) 
due to insufficient guidance for 
capturing spatial dependencies.
This underscores the necessity of 
achieving spatially coherent feature aggregation.

\item 
The classification performance shows a pronounced spatial dependency, 
achieving optimal values near the central token position 
\(c_{255}\). 
Performance decline radially as token positions deviate from the center. 
For example, \(c_{124}\) 
(proximate to the middle class token \(t_{128}\)) 
yields an OA of 92.14\%, AA of 89.16\% and Kappa of 88.39e-2,
indicating a substantial performance degradation
compared to \(c_{256}\).

\item 
Overall, 
spiral scan strategy efficiently enables 
systematic learning of 
spatial dependency  
and classification reasoning 
within the Mamba architecture.

\end{enumerate}

\subsection{Comparisons and Results}\label{Comparisons and Results}
To showcase the advantages of ECP-Mamba, 
we compare it with
4 supervised methods [1)-4)],
2 generative self-supervised methods [5)-6)],
and 2 contrastive self-supervised methods [7)-8)]
on four benchmark PolSAR datasets.
The competitors are detailed below:


\begin{enumerate}[label=\arabic*)]

\item \textbf{RLRMF-RF}~\cite{ref2}: 
A supervised PolSAR image classification method based on random forest classifier and features extracted with 
low-rank matrix factorization (RLRMF)  
to suppress speckle noise.

\item \textbf{RV-CNN}~\cite{zhouyuCNN}: 
The classic PolSAR image classification method based on real-valued CNN.

\item \textbf{CV-CNN}~\cite{baseline_of_fudan_CV_CNN_2017}: 
The benchmark method employing CV-CNN in PolSAR image classification realm.

\item \textbf{PolSF}~\cite{jamali2023local}:
A supervised PolSAR image classification method 
based on local attention-based Transformer.
To ensure its results with limited annotations,
we introduce data augmentation techniques 
including random flipping and rotation.

\item \textbf{3D-GAN}~\cite{jamali2023polsar}:
A 3D generative adversarial network that is capable of generating high-quality synthetic PolSAR data, which overcomes the difficulty of insufficient labels for PolSAR image classification.

\item \textbf{DB-GC}~\cite{wang2024dual}:
A dual-branch PolSAR image classification model based on generative self-supervised learning, which fuses
superpixel-level features learned by graph masked auto-encoder and 
pixel-level features learned by CNN.

\item \textbf{CL-ViT}~\cite{dong2021exploring}: 
A ViT-based PolSAR representation learning framework
with improved self-supervised contrastive learning pattern.

\item \textbf{PiCL}~\cite{kuang2024polarimetry}:
Another state-of-the-art self-supervised PolSAR image classification method, which innovatively integrates polarimetric domain knowledge into the design of contrastive learning framework.

\end{enumerate}

\begin{figure*}[htb]
\begin{center}
\includegraphics[width=1\textwidth]{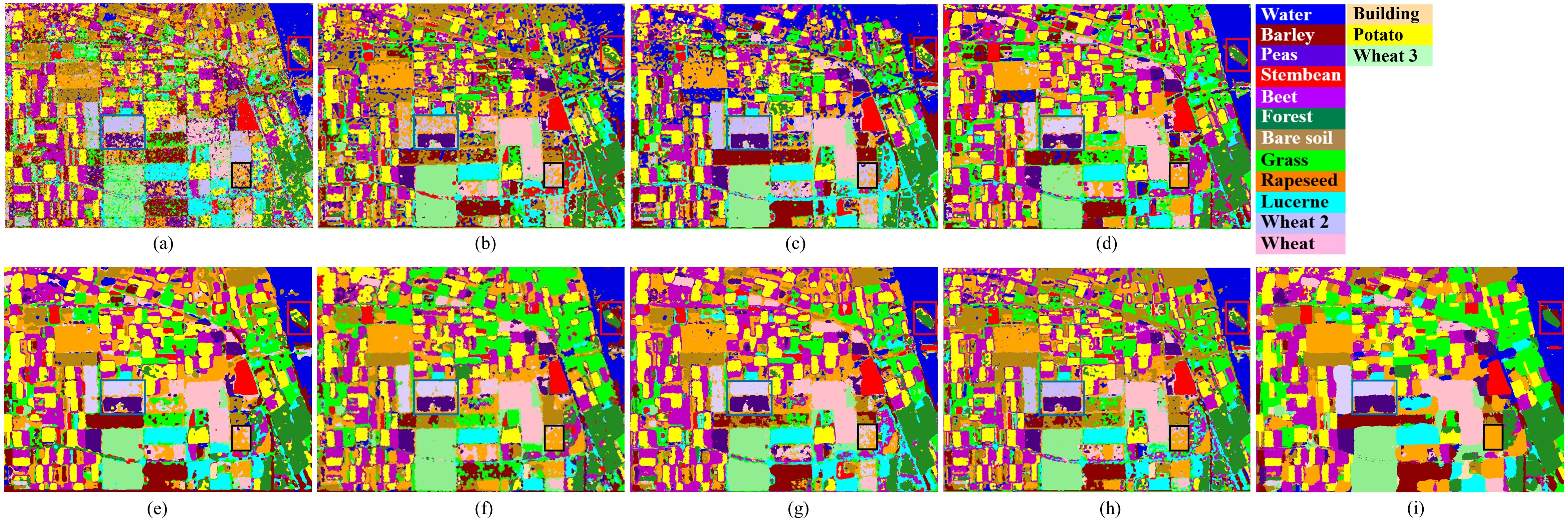}
\vspace{-0.7cm}
\caption{
    Classification results on Flevoland 1989 dataset with different methods.
    (a) RLRMF-RF. 
    (b) RV-CNN.   
    (c) CV-CNN.    
    (d) PolSF.  
    (e) 3D-GAN.    
    (f) DB-GC.    
    (g) CL-ViT.   
    (h) PiCL.  
    (i) ECP-Mamba.
}
\label{Fig_Flevoland1989}
\end{center}
\end{figure*}

\begin{table*}
\caption{
         Comparisons on Flevoland 1989 Dataset with Different Methods
        }
\renewcommand\arraystretch{1.5} 
    \centering
    \resizebox{1.0\linewidth}{!}
    {
    \begin{threeparttable}
        \begin{tabular}{lc ccccccccccc}
        \toprule\hline
        Class & Used labels 
        & RLRMF-RF\cite{ref2} 
        & RV-CNN\cite{zhouyuCNN} 
        & CV-CNN\cite{baseline_of_fudan_CV_CNN_2017}
        & PolSF\cite{jamali2023local}
        & 3D-GAN\cite{jamali2023polsar} 
        & DB-GC\cite{wang2024dual}
        & CL-ViT\cite{dong2021exploring} 
        & PiCL\cite{kuang2024polarimetry}
        & \textbf{Ours} \\
        \midrule
        1: Water & 27 & 95.56 & 86.15 & 87.94 & 99.47 & 96.84 & 93.45 & 99.28 & 97.48 & \textbf{99.84}
        \\
        2: Barley & 16 & 62.94 & 83.65 & 98.83 & 94.58 & 98.59 & 59.10 & 97.38 & 97.77 & \textbf{100}
        \\
        3: Peas & 20 & 63.56 & 95.99 & 97.52 & 92.51 & 97.45 & 97.49 & 98.80 & 96.03 & \textbf{100}
        \\
        4: Stembeans & 13 & 89.90 & 95.31 & 91.86 & 98.52 & 97.96 & 98.67 & 93.72 & 96.53 & \textbf{99.56}
        \\
        5: Beet & 21 & 83.12 & 73.63 & 82.68 & 97.44 & 94.90 & 94.24 & 97.69 & 96.71 & \textbf{99.23}
        \\
        6: Forest & 37 & 88.37 & 93.07 & 93.94 & 97.02& 94.47 & 95.18 & 98.95 & 96.14 & \textbf{99.99}
        \\
        7: Bare soil & 11 & 89.06 & 84.54 & 20.75 & 0 & 97.28 & 98.43 & 96.48 & 99.78 & \textbf{99.90}
        \\
        8: Grass & 15 & 38.86 & 53.30 & 72.63 & 84.16 & 92.45 & 96.06 & 71.07 & 88.48 & \textbf{99.69}
        \\
        9: Rapeseed & 28 & 66.70 & 59.36 & 61.78 & 91.09 & 87.15 & 90.82 & 71.12 & 92.29 & \textbf{99.60}
        \\
        10: Lucerne & 21 & 86.63 & 84.30 & 95.11 & 95.69 & 91.16 & 93.34 & 95.51 & 85.53 & \textbf{99.78}
        \\
        11: Wheat 2 & 23 & 76.60 & 44.80 & 88.88 & 74.89 & 88.94 & 90.90 & 86.35 & 89.60 & \textbf{98.07}
        \\
        12: Wheat & 33 & 70.70 & 89.95 & 93.21 & 91.63 & 94.97 & 93.75 & 98.36 & 98.77 & \textbf{99.76}
        \\
        13: Buildings & 2 & 0.95 & 0 & 0 & 0 & 85.44 & 95.71 & 34.42 & 93.06 & \textbf{99.32}
        \\
        14: Potatoes & 33 & 82.61  & 93.11 & 95.57 & 96.73 & 99.39 & 95.06 & 98.20 & 98.78 & \textbf{99.89}
        \\
        15: Wheat 3 & 45 & 67.32 & 98.54 & 99.16 & 95.71 & 96.25 & 99.09 & 98.32 & 97.61 & \textbf{100}
        \\ \hline
        OA(\%) & - & 75.91 & 82.74 & 87.46 & 90.23& 94.72 & 93.37 & 93.53 & 95.45  & \textbf{99.70}
        \\
        AA(\%) & - & 70.86 & 75.71 & 78.66 & 80.63 & 94.22 & 92.75& 89.04 & 94.97  & \textbf{99.64}
        \\ 
        Kappa(e-2) & - & 68.78 & 73.98 & 77.13 & 79.24& 94.25 & 92.78  & 88.26 & 94.61 & \textbf{99.62}
        \\
        \hline\bottomrule
    \end{tabular}

    \begin{tablenotes}
        \footnotesize 
        \item[*] 
        The best results are shown in bold.
        The `-' in the table means that `Used labels' column does not apply to the 3 assessment criteria.
    \end{tablenotes}
    \end{threeparttable}
    }
\label{Tab_Flevoland1989}
\end{table*}

\begin{figure*}[htb]
\begin{center}
\includegraphics[width=0.89\textwidth]{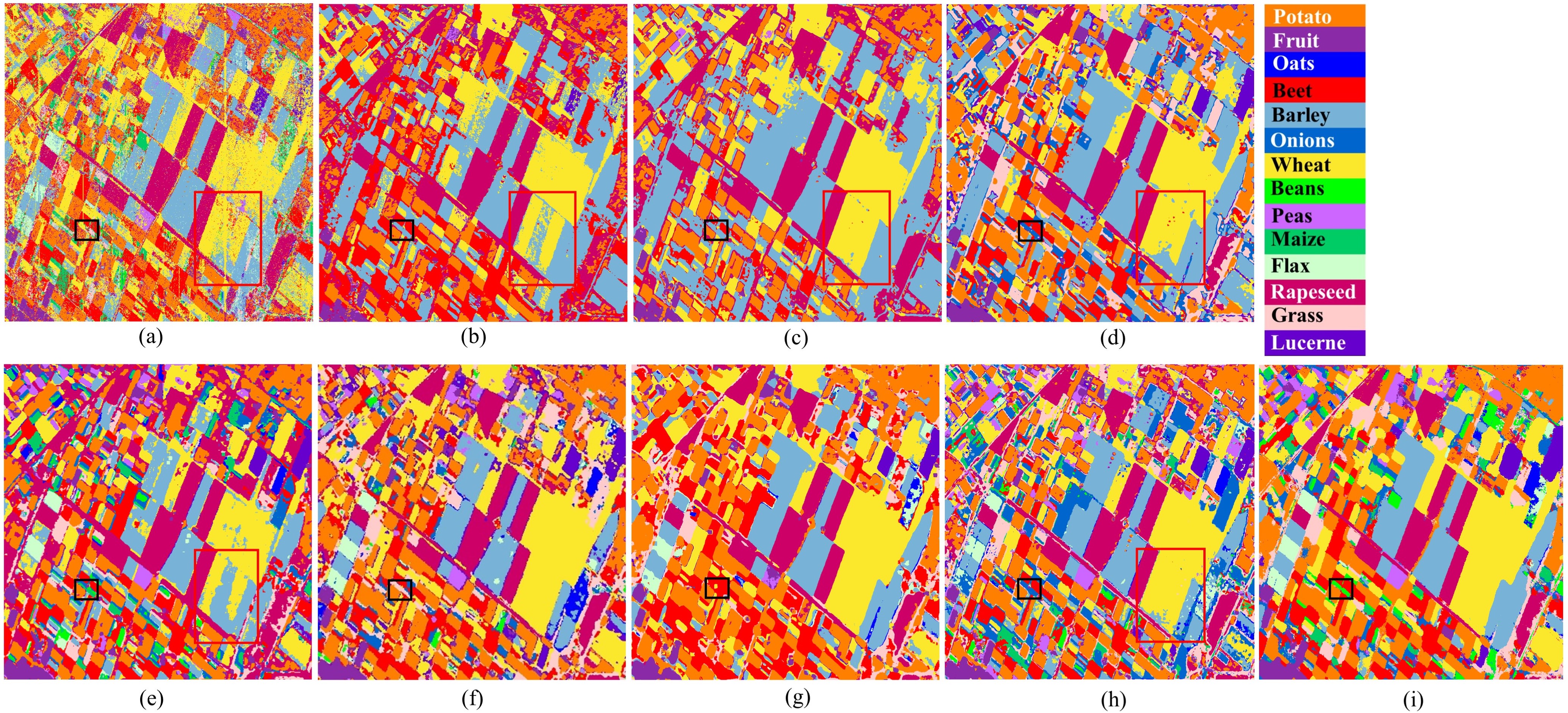}
\vspace{-0.45cm}
\caption{
    Classification results on Flevoland 1991 dataset with different methods.
    (a) RLRMF-RF. 
    (b) RV-CNN.   
    (c) CV-CNN.    
    (d) PolSF.  
    (e) 3D-GAN.    
    (f) DB-GC.    
    (g) CL-ViT.   
    (h) PiCL.  
    (i) ECP-Mamba. 
}
\label{Fig_Flevoland1991}
\end{center}
\end{figure*}

\begin{table*}[htb]
\caption{Comparisons on Flevoland 1991 Dataset with Different Methods}
\renewcommand\arraystretch{1.5}
    \centering
    \resizebox{1.0\linewidth}{!}
    {
    \begin{threeparttable}
        \begin{tabular}{lc ccccccccccc}
        \toprule\hline
        Class & Used labels 
        & RLRMF-RF\cite{ref2} 
        & RV-CNN\cite{zhouyuCNN} 
        & CV-CNN\cite{baseline_of_fudan_CV_CNN_2017}
        & PolSF\cite{jamali2023local}
        & 3D-GAN\cite{jamali2023polsar} 
        & DB-GC\cite{wang2024dual}
        & CL-ViT\cite{dong2021exploring} 
        & PiCL\cite{kuang2024polarimetry}
        & \textbf{Ours} \\
        \midrule
        1: Potato & 44 & 87.39 & 92.67 & 98.45 & 99.83 & 99.68 & 99.71 & 98.50 & 98.95 & \textbf{99.84}
        \\
        2: Fruit & 9 & 83.09 & 92.12 & 88.69 & 99.47& 98.90 & 99.92 & 92.44 & 99.54 & \textbf{100}
        \\
        3: Oats & 3 & 0.14 & 0 & 0 & 0& 98.21 & 92.38 & 6.60 & 98.78  & \textbf{100}
        \\
        4: Beet & 22 & 81.82 & 95.83 & 69.25 & 93.20& 93.61 & 93.77 & 99.10 & 94.65 & \textbf{99.94}
        \\
        5: Barley & 50 & 90.75 & 74.34 & 92.0 & \textbf{99.83} & 91.54 & 93.61 & 98.21 & 92.87 & 99.36
        \\
        6: Onions & 5 & 4.04 & 0 & 0 & 54.69 & 36.29 & 35.83 & 0 & 40.85 & \textbf{54.84}
        \\
        7: Wheat & 53 & 91.80 & 79.31 & 86.25 & 97.71 & 94.65 & 99.23 & 99.14 & 98.87 & \textbf{99.99}
        \\
        8: Beans & 3 & 30.41 & 0 & 0 & 0 & \textbf{80.04} & 55.89 & 9 & 65.25  & 65.71
        \\
        9: Peas & 5 & 69.21 & 8.47 & 0 & 0& 99.72 & 98.11 & 45.74 & 99.54 & \textbf{100}
        \\
        10: Maize & 3 & 58.76 & 0 & 0 & 0& 75.43 & 0 & 0 & \textbf{98.68} & 84.88
        \\
        11: Flax & 9 & 71.61 & 0 & 0 & 0& 99.79 & 100 & 51.92 & 96.91 & \textbf{100}
        \\
        12: Rapeseed & 57 & 93.24 & 91.03 & 99.49 & 99.48 & 92.30 & 98.91 & \textbf{99.90} & 99.60 & 98.84
        \\
        13: Grass & 9 & 45.88 & 0 & 0 & 73.17& 74.38 & 71.51 & 63.39 & 77.33 & \textbf{95.43}
        \\
        14: Lucerne & 6 & 63.82 & 59.86 & 0 & 84.49 & 85.26 & 89.65 & 81.30 & 93.94 & \textbf{99.63}
        \\ \hline
        OA(\%) & - & 83.96 & 74.73 & 78.30 & 89.89& 92.68 & 94.48 & 90.70 & 95.53  & \textbf{98.33}
        \\        
        AA(\%) & - & 62.28 & 42.40 & 38.16 & 57.28& 87.13 & 80.61 & 59.73 & 89.62  & \textbf{92.75}
        \\ 
        Kappa(e-2) & - & 59.38 & 37.97 & 33.40 & 53.99& 91.38 & 93.49 & 56.63 & 87.63  & \textbf{92.19}
        \\
        \hline\bottomrule
    \end{tabular}
    
    \begin{tablenotes}
        \footnotesize 
        \item[*] 
        The best results are shown in bold.
        The `-' in the table means that `Used labels' column does not apply to the 3 assessment criteria.        
    \end{tablenotes}
    
\end{threeparttable}
    }
\label{Tab_Flevoland1991}
\end{table*}

\subsubsection{Flevoland 1989 Dataset}

The classification results of all methods on Flevoland 1989 dataset are shown in Figure~\ref{Fig_Flevoland1989}.
In analyses of the compared methods,
Figure~\ref{Fig_Flevoland1989}(e)-(i) illustrates the classification maps produced by 5 SSL methods. 
These results show a notably closer match to the ground truth
compared to those generated by supervised methods in Figure~\ref{Fig_Flevoland1989}(a)-(d),
highlighting the benefits of utilizing self-supervision signals when annotations are scarce. 
Notably, 
our proposed ECP-Mamba achieves superior contextual consistency and improved discrimination, 
which is demonstrated by subsequent finds:

\begin{enumerate}
\renewcommand{\theenumi}{\alph{enumi}}
\renewcommand{\labelenumi}{\theenumi)}
\item 
As noted by blue rectangles in Figure~\ref{Fig_Flevoland1989}(a)-(i),
Figure~\ref{Fig_Flevoland1989}(i) demonstrates superior performance in distinguishing
adjacent regions with different classes \emph{peas} and \emph{wheat 2},
and shows the best contextual consistency among all methods.

\item 
As noted by black rectangles in Figure~\ref{Fig_Flevoland1989}(a)-(i), 
all competitors have trouble in 
distinguishing two 
highly similar classes, 
i.e., \emph{rapeseed} and \emph{wheat 2} classes,
while our proposed ECP-Mamba is capable of
telling them apart.

\item 
Figure~\ref{Fig_Flevoland1989}(a)-(h)
exhibit a tendency to misclassify pixels of
\emph{forest} class as \emph{potato} class
in the red-boxed island region.
In contrast, 
Figure~\ref{Fig_Flevoland1989}(i) shows more desirable result with higher spatial connectivity and clearer boundaries.
\end{enumerate}

Table~\ref{Tab_Flevoland1989} summarizes numerical comparisons across evaluated methods,
with optimal results highlighted in bold. 
Key observations are outlined below:

\begin{enumerate}
\renewcommand{\theenumi}{\alph{enumi}}
\renewcommand{\labelenumi}{\theenumi)}

\item 
The proposed ECP-Mamba attains state-of-the-art performance, 
which surpasses the second-ranked PiCL method 
by margins of 4.25\%, 4.67\% and 5.01e-2,
underscoring its robustness in classification 
with scare annotations.

\item
PolSF outperforms other supervised methods
but underperforms against all self-supervised ones.
This indicates the notable benefits and research potential of SSL
in label scarce scenes.

\item 
Under the SR of 0.2\%,
only 2 instances are available for \emph{buildings} class. 
Despite this constraint, 
ECP-Mamba achieves 99.32\% accuracy, 
outperforming PiCL by 6.26\%. 
This gracefully demonstrates the efficacy of Multi-scale Efficient Mamba
in extracting discriminative features through contextual dependency modeling.
\end{enumerate}

These results conclusively demonstrate that ECP-Mamba markedly enhances PolSAR image classification in low-label scenarios, 
showcasing its strong potential for practical deployment of 
Spiral Mamba-based remote sensing solutions.

\begin{figure*}[htb]
\begin{center}
\includegraphics[width=0.89\textwidth]{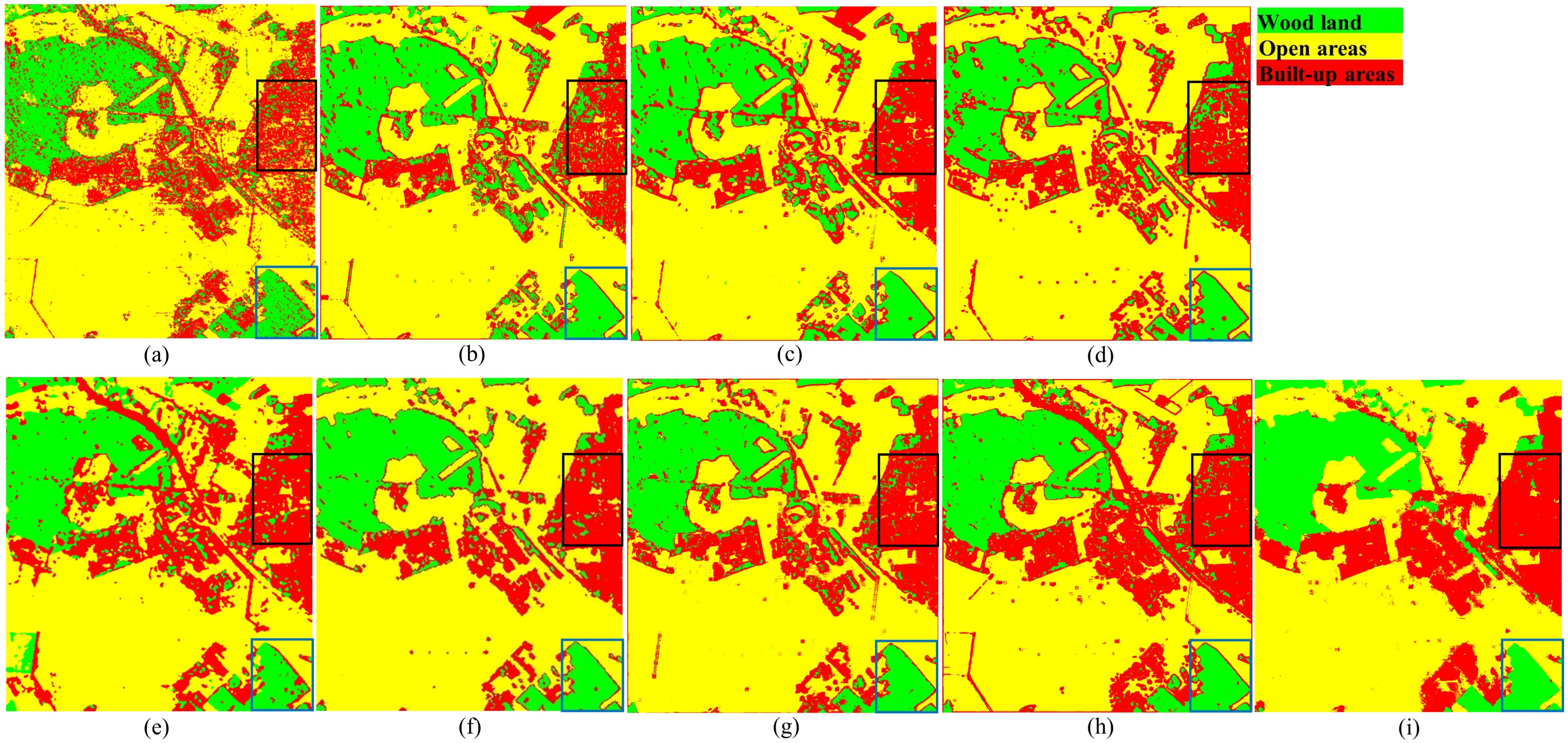}
\vspace{-0.455cm}
\caption{
    Classification results on Oberpfaffenhofen dataset with different methods.        
    (a) RLRMF-RF. 
    (b) RV-CNN.   
    (c) CV-CNN.    
    (d) PolSF.  
    (e) 3D-GAN.    
    (f) DB-GC.    
    (g) CL-ViT.   
    (h) PiCL.  
    (i) ECP-Mamba. 
}
\label{Fig_Oberpfaffenhofen}
\end{center}
\end{figure*}

\begin{table*}[htp]
\caption{Comparisons on Oberpfaffenhofen Dataset with Different Methods}
\renewcommand\arraystretch{1.5}
    \centering
    \resizebox{1\linewidth}{!}
    {
    \begin{threeparttable}
        \begin{tabular}{lc ccccccccccc}
        \toprule\hline
        Class & Used labels 
        & RLRMF-RF\cite{ref2} 
        & RV-CNN\cite{zhouyuCNN} 
        & CV-CNN\cite{baseline_of_fudan_CV_CNN_2017}
        & PolSF\cite{jamali2023local}
        & 3D-GAN\cite{jamali2023polsar} 
        & DB-GC\cite{wang2024dual}
        & CL-ViT\cite{dong2021exploring} 
        & PiCL\cite{kuang2024polarimetry}
        & \textbf{Ours} \\
        \midrule
        1: Built-up Areas & 680 & 69.22 & 62.65 & 69.92 & 73.94  & 76.69 & 76.46 & 77.19 & 83.35 & \textbf{93.06} 
        \\
        2: Wood Land & 537 & 80.72 & 90.31 & 87.76 & 84.65 & 89.70 & 93.32 & 90.13 & 92.22 & \textbf{97.28}
        \\
        3: Open Areas & 1534 & 93.38 & 95.88 & 95.51 & 95.49 & 87.15 & 97.53 & 95.97 & \textbf{97.79} & 97.74
        \\ \hline
        OA(\%) & - & 84.94 & 86.58 & 87.68 & 88.05& 85.06 & 91.51 & 90.19 & 93.13  & \textbf{96.49}
        \\
        AA(\%) & - & 81.11 & 82.95 & 84.40 & 84.69& 84.52 & 89.11 & 87.76 & 91.12  & \textbf{96.03}
        \\ 
        Kappa(e-2) & - & 71.66 & 74.42 & 76.60 & 77.04& 75.24 & 85.38 & 81.64 & 86.68  & \textbf{94.04}
        \\
        \hline\bottomrule
    \end{tabular}
    \begin{tablenotes}
        \footnotesize 
        \item[*] 
        The best results are shown in bold.
        The `-' in the table means that `Used labels' column does not apply to the 3 assessment criteria.        
    \end{tablenotes}
    \end{threeparttable}
   }
\label{Tab_Oberpfaffenhofen}
\end{table*}

\begin{figure*}[htb]
\begin{center}
\includegraphics[width=0.8\textwidth]{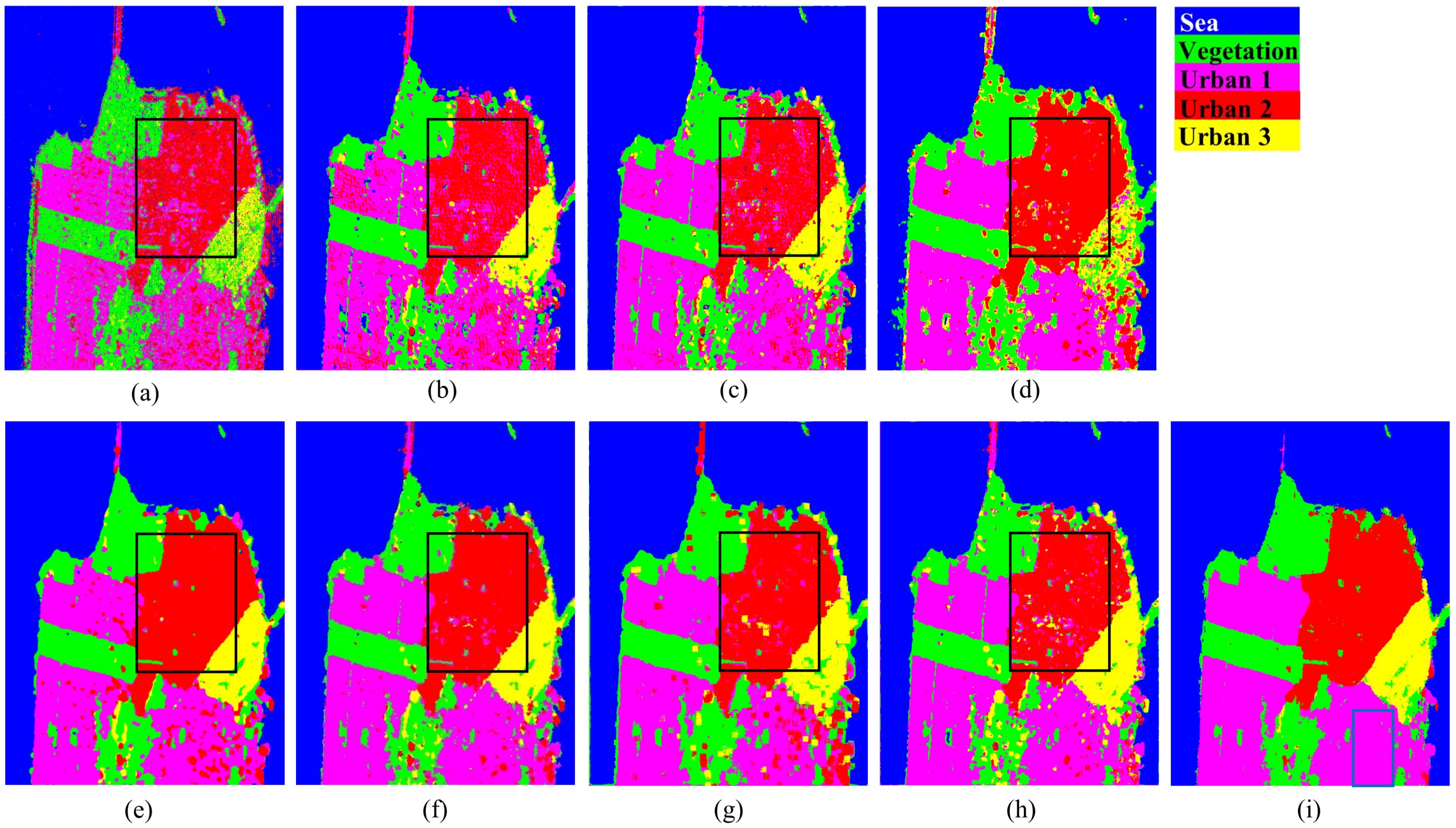}
\vspace{-0.455cm}
\caption{
    Classification results on San Francisco dataset with different methods.    
    (a) RLRMF-RF. 
    (b) RV-CNN.   
    (c) CV-CNN.    
    (d) PolSF.  
    (e) 3D-GAN.    
    (f) DB-GC.    
    (g) CL-ViT.   
    (h) PiCL.  
    (i) ECP-Mamba.
}
\label{Fig_San_Francisco}
\end{center}
\end{figure*}

\begin{table*}[htp]
\caption{
Comparisons on San Francisco Dataset with Different Methods}
\renewcommand\arraystretch{1.5} 
\centering
\resizebox{1.0\linewidth}{!}
{
    \begin{threeparttable}
        \begin{tabular}{lc ccccccccccc}
        \toprule\hline
        Class & Used labels 
        & RLRMF-RF\cite{ref2} 
        & RV-CNN\cite{zhouyuCNN} 
        & CV-CNN\cite{baseline_of_fudan_CV_CNN_2017}
        & PolSF\cite{jamali2023local}
        & 3D-GAN\cite{jamali2023polsar} 
        & DB-GC\cite{wang2024dual}
        & CL-ViT\cite{dong2021exploring} 
        & PiCL\cite{kuang2024polarimetry}
        & \textbf{Ours} \\
        \midrule
        1: Water & 107 & 99.32 & 99.97 & 99.96 & 99.38 & 100 & 99.96 & 98.91 & 99.91 & \textbf{100}
        \\
        2: Vegetation & 30 & 83.21 & 89.40 & 89.62 & 88.59 & 91.55 & 89.37 & 90.97 & 90.95 & \textbf{93.67}
        \\
        3: Urban 1  & 44 & 81.09 & 81.93 & 91.76 & 95.07& 89.53 & 96.14 & 90.65 & 97.48 & \textbf{99.14}
        \\
        4: Urban 2 & 36 & 81.15 & 81.45 & 86.15 & 97.93 & \textbf{99.77} & 97.71 & 95.59 & 96.11 & 99.63
        \\
        5: Urban 3 & 11 & 57.32 & 92.50 & 91.63 & 57.37 & 95.96 & \textbf{96.83} & 90.86 & 95.68 & 95.08
        \\ \hline
        OA(\%) & - & 88.93 & 91.83 & 94.46 & 95.02& 96.63 & 97.33 & 95.38 & 97.48 & \textbf{98.72}
        \\
        AA(\%) & - & 80.42 & 89.05 & 91.82 & 87.67& 95.36 & 96.00 & 93.40 & 96.03 & \textbf{97.50}
        \\ 
        Kappa(e-2) & - & 75.52 & 86.31 & 89.78 & 84.58& 95.16 & 96.16 & 91.75 & 95.04  & \textbf{96.88}
        \\
        \hline\bottomrule
    \end{tabular}
    
    \begin{tablenotes}
        \footnotesize 
        \item[*] 
        The best results are shown in bold.
        The `-' in the table means that `Used labels' column does not apply to the 3 assessment criteria.        
    \end{tablenotes}
\end{threeparttable}
}
\label{Tab_San_Francisco}
\end{table*}

\subsubsection{Flevoland 1991 Dataset}

The classification results for 9 methods on the Flevoland 1991 dataset are shown in Figure~\ref{Fig_Flevoland1991},
with detailed numerical results presented in Table~\ref{Tab_Flevoland1991}.

Figure~\ref{Fig_Flevoland1991} and the ground truth map in Figure~\ref{fig:Experimental_images}(b2)
indicate that ECP-Mamba effectively distinguishes \emph{beans} class in the black-boxed regions,
surpassing other methods. 
As shown in the red box, 
Figure 6(a)-(e) and (h) demonstrate low contextual consistency in \emph{wheat} and \emph{barley} classes. 
In contrast, Figure 6(i) avoids this issue while simultaneously 
achieving more accurate delineation of fine-grained class boundaries.
 
Table~\ref{Tab_Flevoland1991} indicates that ECP-Mamba achieves the best classification results, 
surpassing the second-highest PiCL by 2.8\%, 3.13\% and 4.56e-2 in OA, AA and Kappa values respectively.
This superiority can be attributed to its advanced architecture and training strategy.
ECP-Mamba demonstrates remarkable performance
particularly in classes with sparse labels. 
For example, it attains 95.43\% accuracy in the \emph{grass} class 
and 100\% in the \emph{Oats} class, 
substantially outperforming other supervised methods. 
This underscores the advantages of SSL in 
pixel-level classification when annotations are limited.

\subsubsection{Oberpfaffenhofen Dataset}

Figure~\ref{Fig_Oberpfaffenhofen} shows the classification results for all methods on the Oberpfaffenhofen dataset, 
with numerical results in Table~\ref{Tab_Oberpfaffenhofen}.

In Figure~\ref{Fig_Oberpfaffenhofen}(a)-(h),
the black rectangles highlight widespread pixel blocks of the \emph{wood land} class within the \emph{built-up areas}, 
which is inconsistent with the ground-truth in Figure~\ref{fig:Experimental_images}(c2).
In contrast, 
Figure~\ref{Fig_Oberpfaffenhofen}(i) demonstrates the least confusion 
and the best contextual connectivity in \emph{built-up areas}.
Within the blue rectangle of Figure~\ref{Fig_Oberpfaffenhofen}, 
Figure~\ref{Fig_Oberpfaffenhofen}(a)-(e) exhibit numerous scattered instances 
misclassified as \emph{built-up areas} within \emph{wood land} areas. 
Moreover, the boundaries between \emph{wood land} and \emph{open areas} 
are misclassified as \emph{built-up areas} 
in Figure~\ref{Fig_Oberpfaffenhofen}(a)-(h).
However, our method exhibits better alignment with the ground truth and improved contextual connectivity in counterparts.

In Table~\ref{Tab_Oberpfaffenhofen}, 
it can be observed that on the Oberpfaffenhofen dataset, 
CL-ViT does not significantly outperform PolSF, 
with only marginal improvements of 2.14\%, 3.07\% and 4.6e-2 in OA, AA and Kappa values.
This is primarily attributed to the Oberpfaffenhofen dataset having sufficient annotations for supervised classification,
which limits the potential benefits of pre-training. 
Furthermore, with only 3 classes, 
this dataset allows supervised methods to achieve acceptable AA and Kappa values 
as self-supervised methods (such as DB-GC and PiCL). 
Referring to Table~\ref{Tab_Oberpfaffenhofen}, 
we also find that the \emph{built-up areas} class is more challenging to classify than other classes. 
This may stem from the complex scattering mechanisms of buildings, 
making them challenging to differentiate.
In comparison, 
our approach once again delivers the best classification results,
particularly achieving the highest accuracy of 93.06\% in the \emph{built-up areas} class.
This significantly demonstrates the superiority and robustness of ECP-Mamba.

\subsubsection{San Francisco Dataset}

The experimental results on the San Francisco Dataset \cite{SanFranciscoDataset} 
are visually and numerically presented in Figure \ref{Fig_San_Francisco} and Table \ref{Tab_San_Francisco}
respectively.

Based on Figure \ref{Fig_San_Francisco} and the ground truth map in Figure \ref{fig:Experimental_images}(d2), 
Figure \ref{Fig_San_Francisco}(i) demonstrates higher semantic consistency 
within the black rectangle than other results, 
as numerous isolated pixels in the \emph{urban 2} areas 
are misclassified as \emph{urban 1} and \emph{urban 3} classes
in Figure \ref{Fig_San_Francisco}(a)-(h).
Our method also effectively identifies the \emph{urban 1} class. 
For instance, there is little irregular confusion between \emph{urban 1} and \emph{urban 2} classes
within the blue-boxed area of Figure \ref{Fig_San_Francisco}(h), 
which is a problem evident in the counterparts of Figure \ref{Fig_San_Francisco}(c)-(g). 
These indicate that ECP-Mamba successfully achieves better pixel connectivity and semantic consistency.

Further analysis of Table \ref{Tab_San_Francisco} 
reveals that the \emph{urban 3} class has the least annotations. 
In this case, 
as a traditional DL method,
RLRMF-RF
has a low accuracy of 57.32\% for \emph{urban 3} class. 
Notably, PolSF's precision for the \emph{urban 3} class is also as low as 57.37\%,
which may be because transformer-based supervised methods typically require more training data 
and are less suitable for datasets with class-imbalanced samples. 
In contrast, self-supervised methods produce significantly better results. 
Overall,
ECP-Mamba outperforms all other methods
with the highest OA, AA and Kappa values of 98.72\%, 97.50\% and 96.88e-2 respectively,
wihch are 1.24\%, 1.47\%, and 1.84e-2 higher than those of the second-best competitor, PiCL.

\subsection{Parameter Analysis}\label{Parameter Analysis}

In this section, 
we initially carry out parameter analysis
on the Flevoland 1989 dataset
in Section \ref{parameter_analysis1}, 
exploring diverse network hyper-parameter configurations. 
Subsequently,
a sensitivity analysis is performed on the same dataset
in Section \ref{parameter_analysis2}, 
focusing on how the classification performance varies across different SR.

\subsubsection{Network Parameters}
\label{parameter_analysis1}

\begin{table}[htb]
\centering
\fontsize{8}{10}
\selectfont
\setlength{\tabcolsep}{3pt}
\renewcommand{\arraystretch}{1}
\caption{Network Parameters Analysis of \\ Multi-scale Efficient Mamba on Flevoland 1989 dataset}
    \begin{tabular}{c ccccc ccc}
        \toprule
        \multirow{3}{*}{Type} &
        \multicolumn{5}{c}{Network Parameters} & 
        \multirow{3}{*}{OA(\%)} & 
        \multirow{3}{*}{AA(\%)} & 
        \multirow{3}{*}{Kappa(e-2)} \\
        \cmidrule {2-6} & 
        \makecell[c]{$[M_s,M_t]$} &
        \makecell[c]{$\hat{M}$} &
        \makecell[c]{$[k, K]$} & 
        \makecell[c]{$[\hat{k}, \hat{K}]$} &
        \makecell[c]{$D$} & \\
        \midrule
        
        \textbf{0} & \textbf{[1, 1]} & \textbf{1} & \textbf{[16, 32]} & \textbf{[1, 2]} & \textbf{192} & \textbf{98.58} & \textbf{97.62} & \textbf{97.45} \\
        
        1 & [1, 2] & - & - & - & - & 97.40 & 96.96 & 96.75 \\
        
        2 & [2, 1] & - & - & - & - & 97.06 & 96.03 & 95.75 \\
        
        3 & - & 0 & - & - & - & 98.05 & 97.44 & 97.25 \\
        
        4 & - & 2 & - & - & - & 98.06 & 97.31 & 97.12 \\
        
        5 & - & - & [8, 16] & [1, 2] & - & 95.72 & 94.67 & 94.29 \\
        
        6 & - & - & [16, 64] & [1, 4] & - & 97.09 & 95.84 & 95.54 \\
        
        7 & - & - & - & - & 64 & 97.67 & 91.21 & 90.58 \\
        
        8 & - & - & - & - & 384 & 98.05 & 97.22 & 97.02 \\
        
        \bottomrule
    \end{tabular}
    \begin{tablenotes}
    \footnotesize 
    \item * The `-' in this table means default settings in Type 0.
    \end{tablenotes}
\label{tab:network_sensitivity_analysis}
\end{table}

To investigate the influences of different network 
parameters of Multi-scale Efficient Mamba, 
we conduct experiments
without pre-training
with the following configurations:
$[M_s,M_t]$ is set to [1, 1], [1, 2] and [2, 1]; 
$\hat{M}$ is set to 0, 1, and 2; 
$[k,K]$ is set to [8, 16], [16, 32] and [16, 64], 
with their corresponding $[\hat{k}, \hat{K}]$ set to [1, 2], [1, 2] and [1, 4]; 
$D$ is set to 64, 192 and 384.
Here $M_s$, $M_t$ and $\hat{M}$ are the number of BBPS blocks
in local, global encoder
and Cross Mamba module respectively;
$k$, $K$ are the patch sizes of local and global view;
$\hat{k}$, $\hat{K}$ are the patchify kernel sizes
in local and global branch;
$D$ is the channel dimension.
The classification results are illustrated 
in Table~\ref{tab:network_sensitivity_analysis}.
The best results with default settings in Type 0 are shown in bold.
The ‘-’ in this table denotes 
the adoption of default settings.
From the table,
we derive the following insights:

\begin{enumerate}
\renewcommand{\theenumi}{\alph{enumi}}
\renewcommand{\labelenumi}{\theenumi)}

\item Type 0 achieves superior classification accuracy compared to Type 1 and Type 2, indicating that a symmetrical architecture between local and global branches facilitates local-global feature alignment and interactions.

\item Type 3 shows accuracy degradation compared to Type 0, validating the contribution of Cross Mamba. 
Type 4 also exhibits a performance drop, 
suggesting that excessive layers may induce over-fitting of polarization features in limited label scenarios.
Therefore, it is recommended to use two Mamba-based layers in each branch.

\item Type 0 outperforms Type 5 and Type 6,
demonstrating that $[k,K]$=[16, 32] optimally balances 
the local texture preservation and global context integration.
Smaller local patches ($8 \times 8$) restrict 
the receptive field for capturing polarimetric scattering patterns,
while larger global patches ($64 \times 64$) may introduce 
overly coarse-grained features and excessive noise.

\item 
Compared to Type 7 and Type 8,
Type 0 achieves the peak performance.
Specifically, when $D=64$,
accuracy degrades owing to the limited representation space for polarimetric features.
When $D=384$, there is a slight performance drop
due to the potential over-smoothing 
in high-dimensional manifolds.
In comparison,
$D=192$ balances the computational complexity and classification performance.

\item
Overall,
the best result is achieved with $[M_s,M_t]$, $\hat{M}$, $[k,K]$, $[\hat{k}, \hat{K}]$ and $D$ 
set to [1, 1], 1, [16, 32], [1, 2] and 192, respectively.
Thereinto, these settings are employed as the default settings.

\end{enumerate}

\subsubsection{Sampling Rate}
\label{parameter_analysis2}
\begin{figure}[htb]
    \vspace{-0.1cm}
        \begin{center}
        \hspace*{-0.75cm}
            \includegraphics[width=0.45\textwidth]{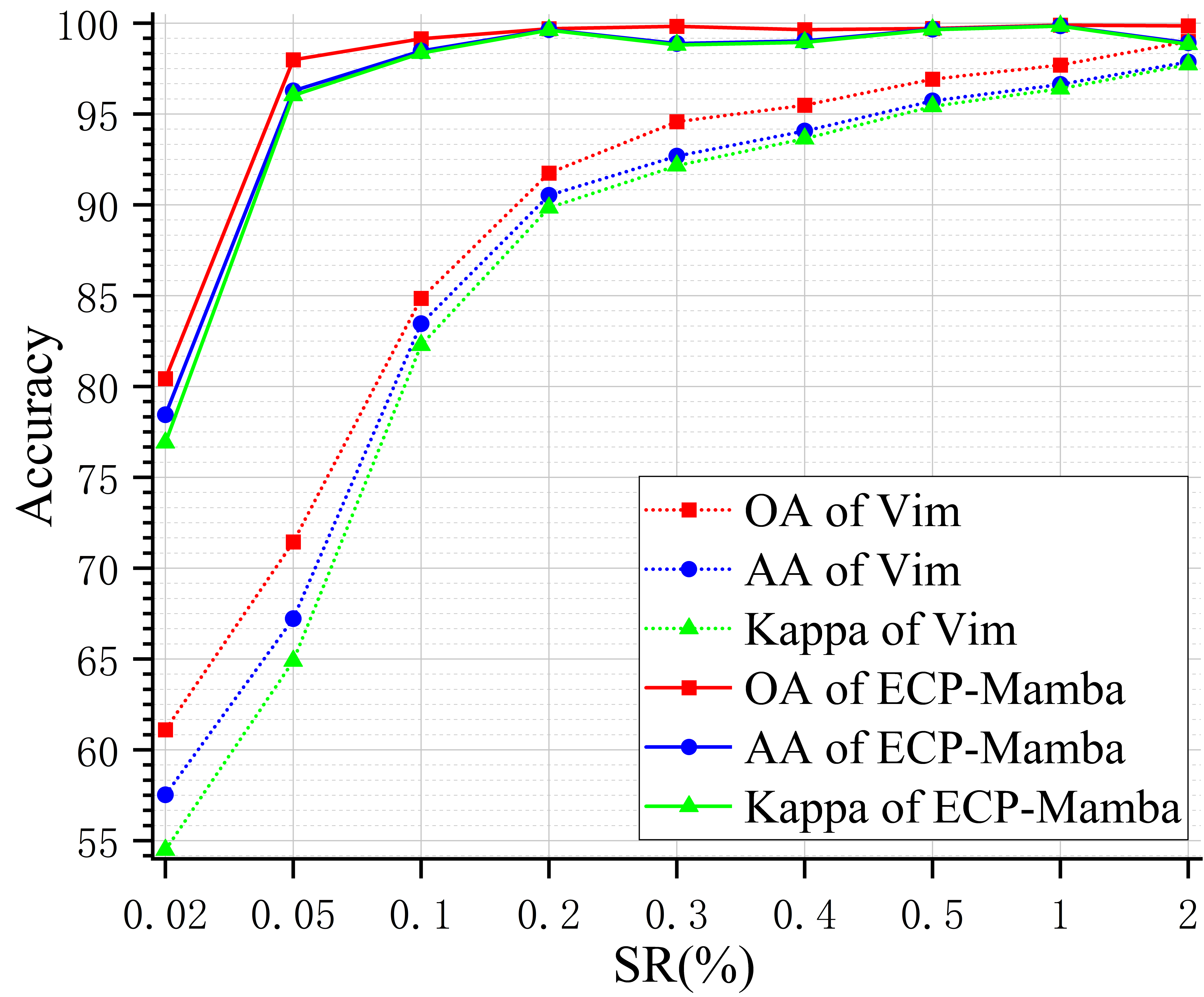}
            \vspace{-0.40cm}
            \caption{Analysis of 
            OA(\%), AA(\%) and Kappa(e-2)
            at various sampling rates (SRs) for 
            the Flevoland 1989 Dataset}
            \label{sampling_rate_sensitivity_analysis}
        \end{center}
\end{figure}

ECP-Mamba undergoes sensitivity analysis across varying SR. 
As depicted in Table \ref{tab:sampling_rate_sensitivity_analysis} and Figure \ref{sampling_rate_sensitivity_analysis}, 
the OA, AA and Kappa values
along with their curves 
are presented for Vim and ECP-Mamba. 
Analysis of figure
combined with numerical results 
reveals the following findings:

\begin{table}[h]
    \centering
    \fontsize{8}{10}
    \selectfont
    \setlength{\tabcolsep}{1.5pt}
    \renewcommand{\arraystretch}{1}
    \caption{Parameter Analysis of Sampling Rate (SR) \\ on Flevoland 1989 dataset}
    \vspace{-0.2cm}
\begin{tabular}{l|ccccccccc}
\bottomrule
\diagbox [width=6em,trim=l] {Acc}{SR(\%)} & 0.02 & 0.05 & 0.1 & 0.2 & 0.3 & 0.4 & 0.5 & 1.0 & 2.0 \\
\hline
\multicolumn{10}{c}{Vim} \\
\hline
OA(\%) & 61.11 & 71.45 & 84.85 & 91.75 & 94.58 & 95.49 & 96.92 & 97.70 & 99.00 \\
AA(\%) & 57.52 & 67.23 & 83.46 & 90.52 & 92.68 & 94.07 & 95.72 & 96.63 & 97.87 \\
Kappa(e-2) & 54.49 & 64.89 & 82.28 & 89.84 & 92.16 & 93.64 & 95.42 & 96.39 & 97.72 \\
\bottomrule
\multicolumn{10}{c}{ECP-Mamba} \\
\hline
OA(\%) & 80.43 & 97.99 & 99.14 & 99.70 & 99.83 & 99.64 & 99.71 & 99.90 & 99.86 \\
AA(\%) & 78.45 & 96.28 & 98.47 & 99.64 & 98.88 & 99.02 & 99.67 & 99.85 & 98.92 \\
Kappa(e-2) & 76.91 & 96.02 & 98.36 & 99.62 & 98.80 & 98.95 & 99.64 & 99.84 & 98.84 \\
\bottomrule

\end{tabular}
\label{tab:sampling_rate_sensitivity_analysis}
\end{table}

\begin{itemize}

\item[$\bullet$] 
ECP-Mamba consistently outperforms Vim across 3 evaluation metrics under every tested SR. 
This demonstrates the robustness and adaptability of 
the proposed framework in diverse data sampling conditions.

\item[$\bullet$] Notably, 
the performance superiority of 
ECP-Mamba becomes particularly pronounced 
when SR$\leq$0.2\%.
For instance, Vim yields suboptimal results when SR is 0.05\%,
achieving an OA of 71.45\%, AA of 67.23\% and Kappa of 64.89e-2.
This suboptimal performance is primarily due to the extreme label sparsity.
At such low SR, 
several classes contain as few as 1 label.
In contrast, 
ECP-Mamba mitigates label scarcity
via multi-scale feature pre-training and 
Multi-scale Efficient Mamba-based feature refinement, 
effectively harnessing class-discriminative information
from massive unlabeled data.

\end{itemize}

\section{Conclusion} 
\label{Conclusion}

This paper introduces ECP-Mamba, 
an efficient multi-scale self-supervised contrastive learning framework 
integrated with state space models
for PolSAR image classification.
ECP-Mamba tackles 2 critical challenges 
in current deep learning-based PolSAR classification:
the reliance on extensive labels and 
the high computational complexity of existing architectures.
The contributions of ECP-Mamba
lie in 3 aspects:
1) A multi-scale self-supervised contrastive learning framework is designed, 
leveraging local-global correspondences 
via a simplified self-distillation paradigm.
2) Through a customized Spiral Mamba,
ECP-Mamba pioneers the use of SSMs 
in PolSAR image classification. 
%
3) A lightweight Cross Mamba module is introduced, 
promoting multi-scale feature interaction 
without additional computational cost.
Extensive experiments on 4 benchmark datasets demonstrate 
ECP-Mamba’s superiority over the state-of-the-art methods
in the challenging scarce label situation.

In the future work,
we plan to focus on 2 directions:
1) We plan to extend this framework to handle heterogeneous PolSAR data distributions
in PolSAR image classification.
2) We are interested in exploring Mamba-based
multi-modal self-supervised learning for 
multi-source remote sensing applications.

\section*{Acknowledgment}
The authors express sincere gratitude to the editor and
anonymous reviewers for their insightful comments and suggestions,
and to the PolSAR data providers for supplying the images.

\ifCLASSOPTIONcaptionsoff
  \newpage
\fi

\bibliographystyle{IEEEtran}

\bibliography{refs} 

\end{document}